\documentclass{article}

    \usepackage[final]{neurips_2022}

\usepackage[utf8]{inputenc} 
\usepackage[T1]{fontenc}    
\usepackage{hyperref}
\hypersetup{colorlinks,citecolor=blue}
\usepackage{url}            
\usepackage{booktabs}       
\usepackage{amsfonts}       
\usepackage{nicefrac}       
\usepackage{microtype}      
\usepackage{xcolor}         
\usepackage{subfigure}
\usepackage{algorithm}
\usepackage{algorithmicx}
\usepackage{algpseudocode}
\usepackage{amsmath}
\usepackage{amsfonts}       
\usepackage{amsthm}
\usepackage{wrapfig}
\usepackage{setspace}
\usepackage{multirow}
\usepackage{tabularx}
\usepackage{xspace}
\usepackage{makecell}
\DeclareMathOperator*{\argmax}{arg\,max}

\DeclareMathOperator*{\sign}{sign}
\newcommand{\eg}{e.g.,\xspace}
\newcommand{\ie}{i.e.,\xspace}
\newcommand{\etal}{et al.\xspace}

\newcommand{\norm}[1]{\left\lVert#1\right\rVert}

\newcommand{\abs}[1]{\left|#1\right|}

\usepackage{lineno}
\usepackage{enumitem}

\newcommand{\revised}[1]{#1}

\title{CalFAT: Calibrated Federated Adversarial Training 
with Label Skewness}

\usepackage{amsmath}
\usepackage{amssymb}
\usepackage{mathtools}
\usepackage{amsthm}

\usepackage[capitalize,noabbrev]{cleveref}

\theoremstyle{plain}

\newtheorem{proposition}{Proposition}
\newtheorem{lemma}{Lemma}

\theoremstyle{definition}
\newtheorem{definition}{Definition}

\theoremstyle{remark}

\algrenewcommand\algorithmicindent{1.2em}

\author{
   Chen Chen\thanks{Work done during internship at Sony AI.} \\
   Zhejiang University \\
   \And
   Yuchen Liu \\
   Zhejiang University \\
  \AND
   Xingjun Ma \\
   Fudan University \\
  \And
   Lingjuan Lyu\thanks{Corresponding author.} \\
   Sony AI \\
}

\begin{document}

\maketitle

\begin{abstract}
Recent studies have shown that, like traditional machine learning, federated learning (FL) is also vulnerable to adversarial attacks.
To improve the adversarial robustness of FL, federated adversarial training (FAT) methods have been proposed to apply adversarial training locally before global aggregation. Although these methods demonstrate promising results on independent identically distributed (IID) data, they suffer from training instability on non-IID data with label skewness, resulting in degraded natural accuracy. This tends to hinder the application of FAT in real-world applications where the label distribution across the clients is often skewed. In this paper, we study the problem of FAT under label skewness, and reveal one root cause of the training instability and natural accuracy degradation issues: skewed labels lead to non-identical class probabilities and heterogeneous local models. We then propose a Calibrated FAT (CalFAT) approach to tackle the instability issue by calibrating the logits adaptively to balance the classes. We show both theoretically and empirically that the optimization of CalFAT leads to homogeneous local models across the clients and better convergence points. Code is available at \href{https://github.com/cc233/CalFAT}{GitHub}. 
\end{abstract}

\section{Introduction}
\label{sec:introduction} 
Federated learning (FL) is a privacy-aware learning paradigm that allows multiple participants (clients) to collaboratively train a global model without sharing their private data~\cite{mcmahan2017communication,tan2022fedproto,zhang2022federated,tan2022federated}. In FL, each client follows the conventional machine learning procedure to train a local model on its own data and periodically uploads the local model updates to a central server for global aggregation. 
However, recent studies have shown that, like conventional machine learning, FL is also vulnerable to well-crafted adversarial examples~\cite{lyu2020privacy,zizzo2020fat,hong2021federated,zhou2021adversarially}, \ie at inference time, attackers can add small, human-perceptible adversarial perturbations to the test examples to fool the global model with high success rates.
This raises security and reliability concerns on the implementation of FL in real-world scenarios where such a vulnerability could cause heavy losses~\cite{yang2019federated}. For example, for cross-silo FL in the biomedical domain, a vulnerable global model may cause misdiagnosis, wrong medical treatments, or even the loss of lives.
Similarly, in financial-based cross-silo FL, the lack of adversarial robustness may lead to huge financial losses. 
It is thus imperative to develop a robust FL method that can train adversarially robust global models resistant to different types of adversarial attacks.

In conventional machine learning, adversarial training (AT) has been shown to be one of the most effective defenses against adversarial attacks~\cite{madry2017towards,zhang2019theoretically,chen2022decision}. 
Since the local training in FL is the same as conventional machine learning, recent works~\cite{zizzo2020fat,hong2021federated,zhou2021adversarially} proposed to perform local AT to improve the adversarial robustness of the global model. These methods in general are known as Federated Adversarial Training (FAT). 
AT has been found to be \emph{more} challenging than standard training~\cite{carmon2019unlabeled,zhang2021geometry,zhang2022towards,zhang2022qekd,chen2022decision}, as it generally requires more training data and larger-capacity models. Moreover, adversarial robustness may even be at odds with accuracy \cite{tsipras2018robustness}, meaning that the increase of robustness may inevitably decrease the natural accuracy (i.e., accuracy on natural test data). As a result, the natural accuracy of AT is much lower than standard training~\cite{croce2020reliable}.
This phenomenon also exists in FL, \ie FAT exhibits slower convergence and lower natural accuracy than standard FL, as mentioned by recent studies~\cite{zizzo2020fat,hong2021federated}.

Arguably, FAT will become more challenging if the data are non-independent and identically distributed (non-IID) across the clients. One typical non-IID setting that commonly exists in real-world applications is skewed label distribution~\cite{li2020fedprox}, where different clients have different label distributions. In this paper, we study the problem of FAT on non-IID data with a particular focus on the challenging skewed label distribution setting (formally defined in Section~\ref{sec:skew_label}).
Under conventional training, Xu \etal~\cite{xu2021robust} have shown that adversarially trained models introduce severe performance disparity across different classes.
And such a disparity will be exacerbated under label skewness, ending up with much worse performance on the minority classes~\cite{wang2021imbalanced}.

\begin{wrapfigure}{h}{.5\textwidth}
  \vspace{-0.5cm}
  \centering
  \includegraphics[width=.45\textwidth]{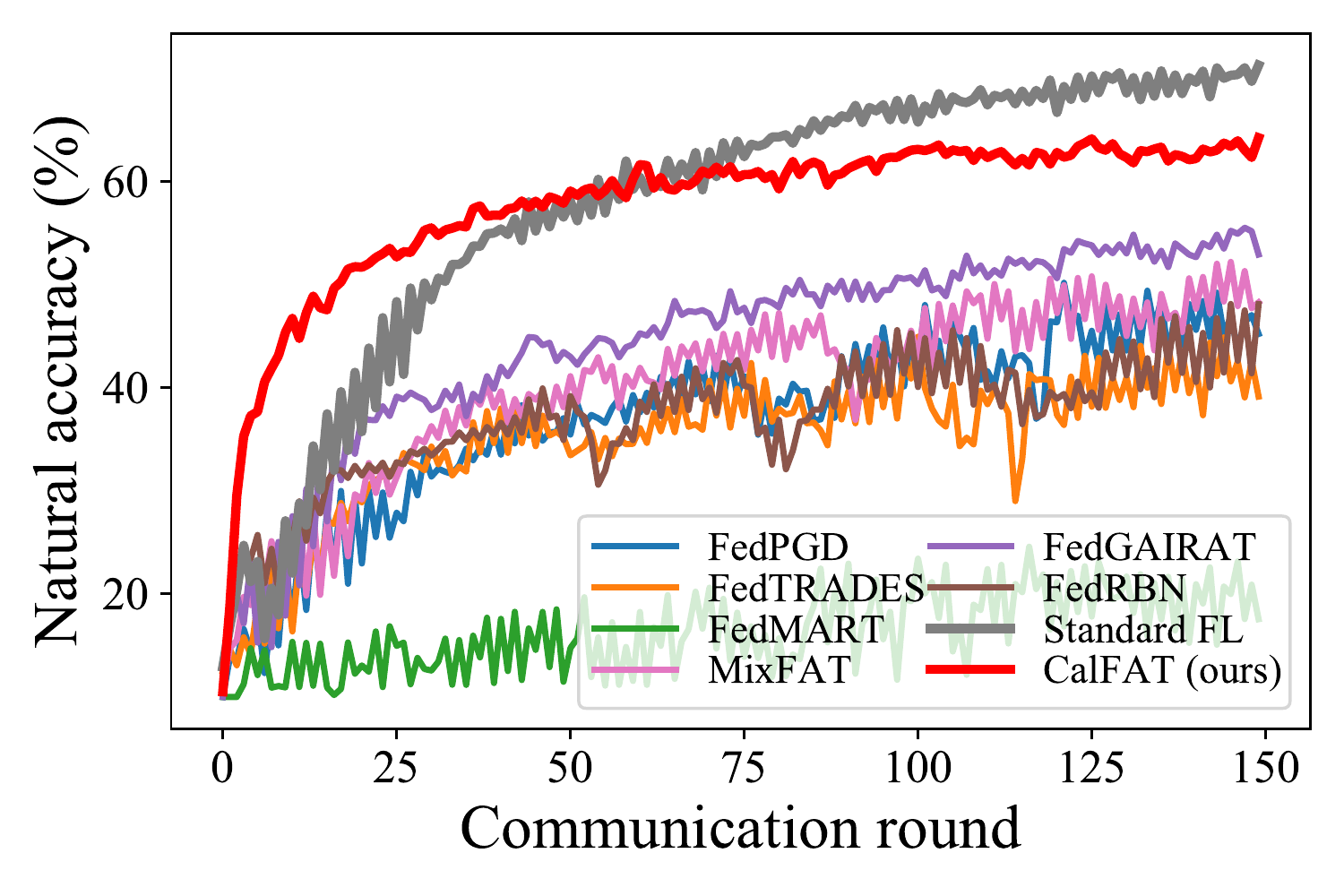}
       \caption{
     Natural accuracy and convergence of standard FL, our CalFAT, and 6 FAT baselines (FedPGD, FedTRADES, FedMART, MixFAT~\cite{zizzo2020fat}, FedGAIRAT and FedRBN~\cite{hong2021federated}) under skewed label distribution with $\beta=0.1$ (see Section~\ref{sec:exp}).
     }
  \label{fig:observation}
\end{wrapfigure}

By far, only a few works have studied non-IID FAT in the current literature.
Zizzo \etal~\cite{zizzo2020fat} propose to perform AT on only part of the local data for better convergence, while standard training is applied to the rest of the local data. We term this method as MixFAT.
Another relevant work called FedRBN~\cite{hong2021federated} tackles a different problem: how to propagate federated robustness to low-resource clients.
Although MixFAT and FedRBN have demonstrated promising results, they suffer from training instability and low natural accuracy issues when compared to standard FL, as we show in Figure~\ref{fig:observation}. We also compare with the other four FAT baselines adapted from existing AT methods to FL, \ie FedPGD, FedTRADES, FedMART, and FedGAIRAT. Unfortunately, these methods also exhibit slow convergence and much degraded final accuracy (details can be found in Section~\ref{sec:exp_results}).
This motivates us to propose a novel method called \emph{Calibrated Federated Adversarial Training} (CalFAT) for effective FAT on non-IID data with skewed label distribution. CalFAT tackles the training instability issue by calibrating the logits to give higher scores to the minority classes.

In summary, our main contributions are:
\begin{itemize}[leftmargin=*]
\item \textit{New insight:} We study the problem of FAT on non-IID data with skewed label distribution, and reveal one root cause of the training instability and natural accuracy degradation: skewed labels lead to non-identical class probabilities and heterogeneous local models.
\item \textit{Novel method:} We propose a novel method called CalFAT for FAT with label skewness, and show that the optimization of CalFAT can lead to homogeneous local models, and consequently, stable training, faster convergence, and better final performance. 
\item \textit{High effectiveness}: Extensive experiments on 4 benchmark vision datasets across various settings prove the effectiveness of our CalFAT and its superiority over existing FAT methods.
\end{itemize}

\section{Notation and Preliminaries}

\subsection{Notation}
Suppose there are $m$ clients in FL with $i$ denoting the $i$-th client, e.g., $\mathcal{D}_i$ denotes the local data of client $i$ and $\theta_i$ denotes the parameters of its local model.
We use $\hat{\theta}$ to denote the parameters of the global model.
Subscript $j$ is the sample index, e.g., $(x_{ij},y_{ij})$ denotes the $j$-th sample of client $i$ and its corresponding label with $y_{ij}\in\{1,\cdots,C\}$.
Let $f_\theta(\cdot)$ be the local model $f(\cdot)$ (before softmax) with parameter $\theta$.
Superscript $l$ is the class index, e.g., $f^l(\cdot)$ denotes the logit output for class $l$.
We denote the adversarial example of clean sample $x$ by $\widetilde{x}$.
$[m]$ denotes the integer set $\{1, \cdots, m\}$.
$p_i(x,y)$ denotes the joint distribution of input $x$ and label $y$ at client $i$, and accordingly, $p_i(y)$ is the marginal distribution of label $y$, $p_i(y\mid x)$ is the conditional distribution of label $y$ given input $x$ and $p_i(x\mid y)$ is the conditional distribution of input $x$ given label $y$.

\subsection{Centralized Adversarial Training}
Let $\mathcal{D}=\{x_{j},y_{j}\}_{j=1}^n$ be the training dataset with $n$ samples.
The cross-entropy loss $\ell_{ce}(f_{\theta}(x),y)$ for an input-label pair $(x,y)$ is defined as $\ell_{ce}(f_{\theta}(x), y)=-\log\sigma^y(f_{\theta}(x))$,
where $\sigma^y(f)=\exp{(f^y)}/\sum_{l=1}^C\exp{(f^l)}$ is the softmax function, $C$ is the number of classes,
and $f^l$ is the model output for class $l$.
The objective function of the centralized adversarial training (AT)~\cite{madry2017towards} can then be defined as $
\min_{\theta}\sum_{j=1}^n\ell_{ce}(f_\theta(\widetilde{x}_j),y_j)/n$,
where the adversarial example $\widetilde{x}_j$ can be generated by
\begin{align}
\widetilde{x}_j=\argmax_{x_j'\in\mathcal{B}_\epsilon(x_j)}\ell_{ce}(f_\theta(x_j'),y_j),
\end{align}
where $\mathcal{B}_\epsilon(x_j)=\{x'\mid \norm{x'-x_j}_{\infty}<\epsilon\}$ is the closed ball of radius $\epsilon>0$ centered at $x_j$, $\norm{\cdot}_\infty$ is the $L_\infty$ norm, and
$\widetilde{x}_j$ is the most adversarial sample within the $\epsilon$-ball. 

A standard centralized AT method uses Projected Gradient Decent (PGD) to generate adversarial examples~\cite{madry2017towards}. 
In particular, PGD iteratively generates an adversarial example $\widetilde{x}_j$ as follows:
\begin{align}
x^{(k+1)}_j=\Pi_{\mathcal{B}_\epsilon(x^{(0)}_j)}\left(x_j^{(k)}+\alpha\sign(\nabla_{x}\ell_{ce}(f_\theta(x_j^{(k)}),y_j))\right),
\quad k=0,\cdots,K-1,
\end{align}
where $k$ is the step number, $K$ is the total number of steps (i.e., $\widetilde{x}_j=x_j^{(K)}$), $\alpha>0$ is the step size, $x^{(0)}_j$ is the natural sample, $x^{(k)}_j$ is the adversarial example generated at step $k$, $\Pi_{\mathcal{B}_\epsilon(x_j^{(0)})}$ is the projection function that projects the adversarial data onto the $\epsilon$-ball centered at $x_j^{(0)}$, and $\sign(\cdot)$ is the sign function. 

By optimizing the model parameters on adversarial examples generated by PGD, centralized AT is able to train a model that is robust against adversarial attacks.

\subsection{Federated Adversarial Training}
\label{sec:fat}
The concept of federated adversarial training (FAT) was first introduced in~\cite{zizzo2020fat} (i.e., MixFAT) to deal with the adversarial vulnerability of FL. MixFAT applies AT locally to improve the robustness of the global model. 
Suppose there are $m$ clients and each client $i$ has its local data $\mathcal{D}_i=\{x_{ij},y_{ij}\}_{j=1}^{n_i}$ sampled from distribution $p_i(x,y)$ with $n_i=\abs{\mathcal{D}_i}$ being the size of the local data.  
In MixFAT, each client $i$ optimizes its local model by minimizing the following objective:
\begin{equation}\label{eq:MixFAT}
\small
\begin{split}
\min_{\theta_i}\frac{1}{n_i}\Big(\sum_{j=1}^{n_i'}\ell_{ce}(f_{\theta_i}(\widetilde{x}_{ij}),y_{ij})+\sum_{j=n_i'+1}^{n_i}\ell_{ce}(f_{\theta_i}(x_{ij}),y_{ij})\Big),
\end{split}
\end{equation}
where 
$\widetilde{x}_{ij}$ is the PGD adversarial example of $x_{ij}$, $n_i'$ is a hyperparameter that controls the ratio of data for AT, and $\theta_i$ are the local model parameters.
After training the local model for certain epochs, client $i$ uploads its local model parameters $\theta_i$ to the central server for aggregation. 
Note that MixFAT only applies AT to a proportion of the local data, mainly for convergence and stability considerations.

\section{Calibrated Federated Adversarial Training (CalFAT)}
\label{sec:calfat}
\subsection{Skewed Label Distribution Leads to Non-identical Class Probabilities}
\label{sec:skew_label}
In this paper, we focus on one representative non-IID setting: \emph{skewed label distribution} \cite{luo2019skeweg,hsieh2020skew}, which is defined as follows.
\begin{definition}[Skewed label distribution]
\label{def:skew}
The label distribution across the clients is skewed,
if for all $i\neq u$ and $i,u\in[m]$:

\quad(a) there exists $y\in[C]$ such that $p_i(y)\ne p_u(y)$\; and \;(b) $p_i(x\mid y)= p_u(x\mid y)$ for all $x, y$.
\end{definition}
Condition (b) is to assume that, given a class $y$, $x$ is sampled with equal probability at different clients.
Note that there exist different types of non-IID: label skew, non-identical class conditional, quantity skew, to name a few (Appendix K in \cite{hsieh2020skew}). The class conditional is often assumed to be identical (i.e., condition (b)) when studying the label skewness problem, which is the main focus of this work. When condition (b) does not hold, it becomes the non-identical class conditional problem.

\begin{lemma}[Non-identical class probabilities]
\label{prop:non_identical_class_prob}
If the label distribution across the clients is skewed and the class conditionals have the same support, then the class probabilities $\{p_i(y\mid x)\mid i\in[m]\}$ are non-identical, \ie{}
for all $i\ne u$ and $i,u\in[m]$, there exist $x$, $y$ such that $p_i(y\mid x)\ne p_u(y\mid x)$.
\end{lemma}

Lemma \ref{prop:non_identical_class_prob} 
implies that skewed label distribution gives rise to non-identical class probabilities $\{p_i(y\mid x)\mid i\in [m]\}$. The proof of Lemma \ref{prop:non_identical_class_prob} is given in Appendix~\ref{appsec:proof_lemma}. 

\subsection{Standard Cross-entropy Leads to Heterogeneity}
\label{sec:heter}

From a statistical point of view, each client $i$ in previous FAT methods estimates its local class probability $p_i(y\mid x)$ during local training \cite{guo2017interpret}.
More specifically, they assume that $p_i(y\mid x)$ can be parameterized by $\theta_i^*$ as:
\begin{align}
\label{eq:parameterize_naivefat}
    p_i(y\mid x)
    =\hat{p}(y\mid x;\theta_i^*)
    =\sigma^y(f_{\theta_i^*}(x)),
\end{align}
where $\theta_i^*$ is the ground-truth parameters of the local class probability $p_i(y\mid x)$.
According to Lemma~\ref{prop:non_identical_class_prob}, the class probabilities $\{p_i(y\mid x)\}$ are non-identical when there is a skewed label distribution.
Therefore, the ground-truth parameters $\{\theta_i^*\mid i\in[m]\}$ are heterogeneous.
We use the sample variance of the ground-truth parameters to measure such heterogeneity as follows:
\begin{align}
\small
    (s^*)^2 =V(\theta_1^*,\ldots,\theta_m^*)= \frac{1}{m-1}\sum_{i=1}^m\|\theta_i^*-\frac{1}{m}\sum_{j=1}^m\theta_j^*\|^2.
\end{align}

Each client $i$ updates its local model parameters $\theta_i$ by optimizing the standard cross-entropy (CE) loss. The updated $\theta_i$ is the maximum likelihood estimate \cite{bickel2015statistics} of the ground-truth parameter $\theta_i^*$ \cite{guo2017interpret}.
Similarly, we use the sample variance \cite{bickel2015statistics} of the local model parameters to measure the heterogeneity of the local models:
\begin{align}
    s^2 = V(\theta_1,\ldots,\theta_m).
\end{align}
Larger sample variance implies higher model heterogeneity. 

The following proposition suggests that the heterogeneity of local models originates from the heterogeneity of the local class probabilities.
\begin{proposition}[Heterogeneous local models]
\label{prop:hetero}
Assume the label distribution across the clients is skewed.
Let $\theta_i$ be the maximum likelihood estimate of $\theta_i^*$ in Eq. \eqref{eq:parameterize_naivefat} given local data at client $i$. 
Then $s^2$ converges almost surely to a nonzero constant:
\begin{align}
    s^2\xrightarrow[]{\text{a.s.}} (s^*)^2\ne 0,
\end{align}
where 
$\xrightarrow[]{\text{a.s.}}$ represents the almost sure convergence.
\end{proposition}
The proof of Proposition \ref{prop:hetero} is provided in Appendix~\ref{appsec:proof_pro1}.
$(s^*)^2$ measures the heterogeneity of the ground-truth parameters $\{\theta_i^*\mid i\in[m]\}$, which reflects the class probability difference across the clients as shown in Eq. \eqref{eq:parameterize_naivefat}.

Proposition \ref{prop:hetero} implies that the local models in previous FAT methods are heterogeneous when the label distribution across the clients is skewed. 
Since the local models are heterogeneous, aggregating these models tends to hurt the convergence and cause the divergence of the global model \cite{li2020fedprox}.
As shown in Figure~\ref{fig:observation}, the training process of existing FAT methods is unstable and has much lower natural accuracy than the standard FL.



\begin{algorithm}[htbp]
\caption{Local training of CalFAT}
\label{alg:calfat}
\textbf{Input:} 

Client $i$, global model parameters $\hat{\theta}$, local dataset $\mathcal{D}_i$, local epoch number $E$, and positive constant $\delta$
\begin{algorithmic}[1]
\setlength{\itemindent}{0em}
\Procedure{ClientUpdate}{}
    \State
    $\theta_i\leftarrow\hat{\theta}$
    \State Compute $\pi_i$ with $\mathcal{D}_i$ by $\pi_i^y=n_i^y/n_i + \delta, y\in[C]$
    \For{local epoch=$1,\cdots,E$ }
    \For{$j=1,\cdots,n_i$}
    \State Sample $(x_{ij},y_{ij})$ from $\mathcal{D}_i$
    \State Generate adversarial example $\widetilde{x}_{ij}=\argmax_{x'_{ij}\in\mathcal{B}_\epsilon(x_{ij})}\ell_{ckl}(f_{\theta_i}(x'_{ij}),f_{\theta_i}(x_{ij}),\pi_i)$
    \EndFor
    \State $\theta_i\leftarrow\theta_i-\eta\frac{1}{n_i}\sum_{j=1}^{n_i}\nabla_{\theta_i}\ell_{cce}(f_{\theta_i}(\widetilde{x}_{ij}),y_{ij},\pi_i)$
    \EndFor
    \State \textbf{return} $\theta_i$
\EndProcedure
\end{algorithmic}
\end{algorithm}

\subsection{Learning Homogeneous Local Models by Calibration}
\label{subsec:homo}
Motivated by \cite{menon2020long},
we propose to re-parameterize the class probabilities.
According to Bayes' formula \cite{klenke2013bayes},
\begin{align}
    p_i(y\mid x)
    =\frac{p_i(x\mid y)p_i(y)}{\sum_{l=1}^C p_i(x\mid l)p_i(l)}.
\end{align}
On the right-hand side of the above equation: (1) the class priors can be easily computed by the relative frequencies \cite{bickel2015statistics};
and (2) more importantly, the class conditionals $\{p_i(x\mid y)\mid i\in[m]\}$ are \emph{identical} across different clients (see Definition~\ref{def:skew}).

Inspired by the above observation, we propose an alternative parameterization of $p_i(y\mid x)$. 
Assume that for all $i\in[m]$, the class conditional $p_i(x\mid y)$ can be parameterized by $\theta^*$ as $
    p_i(x\mid y)=\hat{q}(x\mid y;\theta^*)$,
where $\hat{q}(x\mid y;\theta^*)$ can be an arbitrary conditional probability function.
Then, $p_i(y\mid x)$ can be re-parameterized by $\theta^*$ as follows:
\begin{align}
\label{eq:repara}
    p_i(y\mid x)
    =&{}\hat{q}_i(y\mid x;\theta^*)
    =\frac{\hat{q}(x\mid y;\theta^*)\pi_i^y}{\sum_{l=1}^C\hat{q}(x\mid l;\theta^*)\pi_i^l}.
\end{align}
where 
\begin{align}
\label{eq:pi}
    \pi_i^y=n_i^y/n_i + \delta, y\in[C].
\end{align}
Here $\pi_i^y$ approximates the class prior $p_i(y)$, $n_i^y$ is the sample size of class $y$ on client $i$ and
$\delta>0$ is a small constant added for numerical stability purpose.
During local updates, client $i$ uses its local data to update $\theta_i$, which makes $\theta_i$ the maximum likelihood estimate of $\theta^*$.
The entire training procedure of our method is described in Section~\ref{sec:calfat_detail}.

The following proposition suggests that the local models are homogeneous when trained with the above re-parameterization. The proof of Proposition \ref{prop:homo} is provided in Appendix~\ref{appsec:proof_pro2}.
\begin{proposition}[Homogeneous local models]
\label{prop:homo}
Assume the label distribution across the clients is skewed.
Let $\theta_i$ be the maximum likelihood estimate of $\theta^*$ in Eq. \eqref{eq:repara} given local data at client $i$. 
Then $s^2$ converges almost surely to zero:
\begin{align}
    s^2\xrightarrow[]{\text{a.s.}} 0.
\end{align}
\end{proposition}

\subsection{Details of CalFAT}
\label{sec:calfat_detail}
The local training procedure of our proposed CalFAT is described in Algorithm~\ref{alg:calfat}.
Specifically, we define $\hat{q}(x\mid y;\theta^*)=\exp{(f_{\theta^*}^y(x))}$.
Then, we maximize the likelihood of $\hat{q}_i(y\mid x;\theta^*)$ for each client $i$, which is equivalent to minimizing the following objective:
\begin{equation}\label{eq:fat}
    \begin{split}
    \min_{\theta_i}\frac{1}{n_i}\sum_{j=1}^{n_i}\ell_{cce}(f_{\theta_i}(\widetilde{x}_{ij}),y_{ij},\pi_i),
    \end{split}
\end{equation}
where $\ell_{cce}(\cdot,\cdot,\cdot)$ is the calibrated cross-entropy (CCE) loss and $\widetilde{x}_{ij}$ is the adversarial example of $x_{ij}$. The CCE loss is defined as:
\begin{align}
\label{eq:ce_ours}
    \ell_{cce}(f_{\theta_i}(\widetilde{x}_{ij}), y_{ij},\pi_i){}
    =-\log{\sigma^{y_{ij}}(f_{\theta_i}(\widetilde{x}_{ij})+\log{\pi_i})}.
\end{align}
As discussed in Section \ref{subsec:homo}, minimizing the above CCE loss mitigates the heterogeneity of the local models, which can lead to improved convergence and performance of the global model.

In previous FAT methods, heterogeneous local models tend to give higher scores to the majority classes while lower scores to the minority classes. By contrast, our CalFAT encourages local models to give higher scores to the minority classes by adding a class-wise prior $\log{\pi_i^l}$ to the logits.
Also different from MixFAT that trains the local models on both natural and adversarial data, our CalFAT trains the local models only on  adversarial examples. 
Extensive empirical experiments are conducted in Section~\ref{sec:exp_results} to show the impact of using only adversarial data for optimization.

\paragraph{Adversarial example generation.}
Inspired by~\cite{zhang2019theoretically}, we generate the adversarial examples by maximizing the following calibrated Kullback–Leibler (CKL) divergence loss:
\begin{equation}\label{eq:our_adv_x}
\begin{split}
\widetilde{x}_{ij}=\argmax_{x'_{ij}\in\mathcal{B}_\epsilon(x_{ij})}\ell_{ckl}(f_{\theta_i}(x'_{ij}),f_{\theta_i}(x_{ij}),\pi_i),
\end{split}
\end{equation}
where $\ell_{ckl}(\cdot,\cdot,\cdot)$ is the 
CKL loss defined as:
\begin{equation}\label{eq:kl_ours}
\begin{split}
    \ell_{ckl}(f_{\theta_i}(x'_{ij}),f_{\theta_i}(x_{ij}),\pi_i)
    ={}&-\sum_{y=1}^C\sigma^y(f_{\theta_i}(x_{ij})+\log{\pi_i})\log{\sigma^y(f_{\theta_i}(x'_{ij})+\log{\pi_i})},
\end{split}
\end{equation}
where $\log{\pi_i}$ is the same as in our CCE loss. 
Following centralized AT~\cite{madry2017towards}, we also use PGD to solve Eq.~\eqref{eq:our_adv_x}.


After training the local model for certain epochs following the above procedure, each client $i$ uploads the model parameters $\theta_i$ to the server for aggregation. 
To be consistent with the most recent FAT methods~\cite{zizzo2020fat,hong2021federated}, we adopt the most widely used FedAvg~\cite{mcmahan2017communication} as the default aggregation framework. Our method is compatible with other FL frameworks (\eg FedProx~\cite{li2018federated} and Scaffold~\cite{karimireddy2020scaffold}), as we will show in Section~\ref{sec:exp_results}.

\section{Experiments}
\label{sec:exp}
\paragraph{Data configurations.} Our experiments are conducted on 4 real-world datasets: CIFAR10~\cite{krizhevsky2009learning}, CIFAR100~\cite{krizhevsky2009learning}, SVHN~\cite{netzer2011reading_SVHN}, and ImageNet subset~\cite{deng2009imagenet}.
To simulate label skewness, we 
sample $p_i^{l} \sim Dir(\beta)$ and allocate a $p^l_i$ proportion of the data of label $l$ to client $i$, where $Dir(\beta)$ is the Dirichlet distribution with a concentration parameter $\beta$~\cite{yurochkin2019bayesian}.
By default, we set $\beta=0.1$ to simulate a highly skewed label distribution that widely exists in reality.

\paragraph{Baselines.} We compare our proposed CalFAT with two state-of-the-art FAT methods: MixFAT~\cite{zizzo2020fat} and FedRBN~\cite{hong2021federated}.
We also 
investigate the combination of the state-of-the-art centralized AT methods with FL, i.e., we apply standard PGD~\cite{madry2017towards}, TRADES~\cite{zhang2019theoretically}, MART~\cite{wang2020improving_MART}),
and GAIRAT~\cite{zhang2021geometry}
to FL, and term them as FedPGD, FedTRADES, FedMART, and FedGAIRAT, respectively. 

\paragraph{Evaluation metrics.} We report the natural test accuracy (Natural) and robust test accuracy under 
the most representative attacks, \ie FGSM~\cite{wong2020fast_zico_kolter}, BIM~\cite{kurakin2016adversarial}, PGD-20~\cite{madry2017towards}, CW~\cite{carlini2017towards}, and AA~\cite{croce2020reliable}. 
We run the experiment for 5 times and report the mean and standard deviation.
More detailed experimental setup is provided in Appendix~\ref{appsec:exp_setting}.

\begin{table*}[t]
\caption{Natural and robust accuracy (\%) on different datasets. The best results are in \textbf{bold}.
}
\label{tbl:dataset}
\centering
\resizebox{\textwidth}{!}{
\begin{tabular}{c|cccccc|cccccc}
\toprule
Dataset & \multicolumn{6}{c|}{CIFAR10} & \multicolumn{6}{c}{CIFAR100} \\
\midrule
Metric & Natural & FGSM & BIM & CW & PGD-20 & AA & Natural & FGSM & BIM & CW & PGD-20 & AA\\
\midrule
MixFAT & 53.35	$\pm$ 0.11 & 29.14	$\pm$ 0.10 & 26.31	$\pm$ 0.17 & 22.79	$\pm$ 0.12 & 26.27	$\pm$ 0.11 & 21.89	$\pm$ 0.13 & 34.43	$\pm$ 0.13 & 15.69	$\pm$ 0.13 & 14.60	$\pm$ 0.14 & 11.31	$\pm$ 0.17 & 14.36	$\pm$ 0.20 & 9.06	$\pm$ 0.11 \\
FedPGD & 46.96	$\pm$ 0.16 & 28.70	$\pm$ 0.19 & 26.59	$\pm$ 0.18 & 24.38	$\pm$ 0.17 & 26.74	$\pm$ 0.18 & 22.47	$\pm$ 0.11 & 33.96	$\pm$ 0.14 & 16.07	$\pm$ 0.08 & 14.68	$\pm$ 0.10 & 11.67	$\pm$ 0.10 & 14.67	$\pm$ 0.15 & 10.87	$\pm$ 0.12 \\
FedTRADES & 46.06	$\pm$ 0.12 & 27.75	$\pm$ 0.17 & 26.32	$\pm$ 0.09 & 22.86	$\pm$ 0.10 & 26.31	$\pm$ 0.12 & 21.70	$\pm$ 0.09 & 29.55	$\pm$ 0.10 & 15.01	$\pm$ 0.06 & 14.11	$\pm$ 0.11 & 10.58	$\pm$ 0.03 & 14.30	$\pm$ 0.13 & 9.53	$\pm$ 0.09 \\
FedMART & 25.67	$\pm$ 0.21 & 18.50	$\pm$ 0.18 & 18.21	$\pm$ 0.22 & 15.22	$\pm$ 0.17 & 18.10	$\pm$ 0.22 & 14.41	$\pm$ 0.20 & 19.96	$\pm$ 0.17 & 13.00	$\pm$ 0.19 & 12.91	$\pm$ 0.14 & 9.92	$\pm$ 0.21 & 12.83	$\pm$ 0.18 & 8.57	$\pm$ 0.14 \\
FedGAIRAT & 48.42	$\pm$ 0.08 & 29.30	$\pm$ 0.09 & 26.55	$\pm$ 0.07 & 22.78	$\pm$ 0.12 & 27.20	$\pm$ 0.08 & 21.96	$\pm$ 0.07 & 34.92	$\pm$ 0.05 & 16.18	$\pm$ 0.06 & 15.37	$\pm$ 0.10 & 11.80	$\pm$ 0.05 & 14.90	$\pm$ 0.03 & 9.41	$\pm$ 0.05 \\
FedRBN & 47.80	$\pm$ 0.06 & 26.87	$\pm$ 0.07 & 26.25	$\pm$ 0.03 & 22.00	$\pm$ 0.01 & 26.30	$\pm$ 0.09 & 21.33	$\pm$ 0.09 & 28.55	$\pm$ 0.07 & 14.69	$\pm$ 0.04 & 13.41	$\pm$ 0.08 & 9.71	$\pm$ 0.08 & 14.15	$\pm$ 0.12 & 8.83	$\pm$ 0.08 \\
CalFAT (ours) & \textbf{64.69}	$\pm$ 0.08 & \textbf{35.03}	$\pm$ 0.12 & \textbf{31.50}	$\pm$ 0.07 & \textbf{24.69}	$\pm$ 0.11 & \textbf{31.12}	$\pm$ 0.11 & \textbf{22.91}	$\pm$ 0.08 & \textbf{44.57}	$\pm$ 0.10 & \textbf{17.63}	$\pm$ 0.10 & \textbf{15.60}	$\pm$ 0.11 & \textbf{12.01}	$\pm$ 0.11 & \textbf{15.21}	$\pm$ 0.07 & \textbf{11.49}	$\pm$ 0.08 \\
\midrule
Dataset & \multicolumn{6}{c|}{SVHN} & \multicolumn{6}{c}{ImageNet subset} \\
\midrule
Metric & Natural & FGSM & BIM & CW & PGD-20 & AA & Natural & FGSM & BIM & CW & PGD-20 & AA\\
\midrule												
MixFAT & 19.57	$\pm$ 0.10 & 19.61	$\pm$ 0.12 & 19.66	$\pm$ 0.12 & 19.66	$\pm$ 0.11 & 19.75	$\pm$ 0.11 & 14.80	$\pm$ 0.07 & 33.53	$\pm$ 0.06 & 19.47	$\pm$ 0.02 & 18.48	$\pm$ 0.10 & 16.15	$\pm$ 0.07 & 18.39	$\pm$ 0.02 & 11.98	$\pm$ 0.06 \\
FedPGD & 19.55	$\pm$ 0.08 & 19.33	$\pm$ 0.09 & 19.37	$\pm$ 0.08 & 19.68	$\pm$ 0.05 & 19.52	$\pm$ 0.09 & 13.64	$\pm$ 0.10 & 30.87	$\pm$ 0.12 & 18.88	$\pm$ 0.13 & 17.95	$\pm$ 0.10 & 16.07	$\pm$ 0.11 & 18.40	$\pm$ 0.16 & 11.34	$\pm$ 0.08 \\
FedTRADES & 56.96	$\pm$ 0.13 & 36.92	$\pm$ 0.13 & 35.15	$\pm$ 0.05 & 31.08	$\pm$ 0.14 & 34.90	$\pm$ 0.15 & 30.37	$\pm$ 0.11 & 30.22	$\pm$ 0.14 & 18.67	$\pm$ 0.13 & 17.99	$\pm$ 0.21 & 16.23	$\pm$ 0.13 & 17.82	$\pm$ 0.12 & 11.81	$\pm$ 0.10 \\
FedMART & 19.85	$\pm$ 0.16 & 19.94	$\pm$ 0.16 & 19.71	$\pm$ 0.16 & 19.85	$\pm$ 0.15 & 19.79	$\pm$ 0.17 & 14.64	$\pm$ 0.14 & 26.47	$\pm$ 0.18 & 16.40	$\pm$ 0.18 & 15.53	$\pm$ 0.17 & 14.43	$\pm$ 0.13 & 15.40	$\pm$ 0.21 & 9.34	$\pm$ 0.14 \\
FedGAIRAT & 58.41	$\pm$ 0.11 & 38.30	$\pm$ 0.12 & 36.52	$\pm$ 0.09 & 31.24	$\pm$ 0.15 & 36.69	$\pm$ 0.13 & 31.63	$\pm$ 0.06 & 34.25	$\pm$ 0.12 & 19.62	$\pm$ 0.10 & 19.28	$\pm$ 0.14 & 16.78	$\pm$ 0.14 & 19.18	$\pm$ 0.12 & 11.80	$\pm$ 0.09 \\
FedRBN & 53.88	$\pm$ 0.04 & 34.48	$\pm$ 0.08 & 32.52	$\pm$ 0.02 & 27.99	$\pm$ 0.02 & 32.32	$\pm$ 0.03 & 28.35	$\pm$ 0.05 & 29.35	$\pm$ 0.09 & 18.76	$\pm$ 0.03 & 17.25	$\pm$ 0.12 & 15.07	$\pm$ 0.10 & 18.05	$\pm$ 0.09 & 11.42	$\pm$ 0.12 \\
CalFAT (ours) & \textbf{84.15}	$\pm$ 0.07 & \textbf{48.38}	$\pm$ 0.11 & \textbf{42.04}	$\pm$ 0.07 & \textbf{31.66}	$\pm$ 0.04 & \textbf{41.68}	$\pm$ 0.11 & \textbf{32.57}	$\pm$ 0.10 & \textbf{49.89}	$\pm$ 0.11 & \textbf{22.31}	$\pm$ 0.17 & \textbf{19.99}	$\pm$ 0.09 & \textbf{17.42}	$\pm$ 0.12 & \textbf{19.97}	$\pm$ 0.14 & \textbf{12.30}	$\pm$ 0.07 \\
\bottomrule
\end{tabular}}
\end{table*}

\subsection{Main Results}
\label{sec:exp_results}

\paragraph{Evaluation on different datasets.}
Table~\ref{tbl:dataset} shows the results of all methods on CIFAR10, CIFAR100, SVHN, and ImageNet subset. From the table, we can observe that:

(1) Our CalFAT achieves the best robustness on all datasets, validating the efficacy of our CalFAT. 
For example, CalFAT outperforms the best baseline method (FedGAIRAT) by 10.20\% on SVHN dataset under FGSM attack. 

(2) Our CalFAT shows a 
significant improvement in natural accuracy compared to other baselines. For example, CalFAT can improve the natural accuracy of the best baseline method (FedGAIRAT) by 25.63\% on SVHN dataset. We hypothesise that the reason lies in 
the homogeneity of local models in our CalFAT, which leads to better convergence and higher clean accuracy.

(3) All methods demonstrate the worst performance on CIFAR100 and ImageNet subset datasets. We conjecture that this is because there are more classes in these two datasets, making federated training substantially harder. Nevertheless, our CalFAT still achieves the best performance.




\begin{figure}[t]
\begin{minipage}{0.35\textwidth}
\centering
\includegraphics[height=2.9cm]{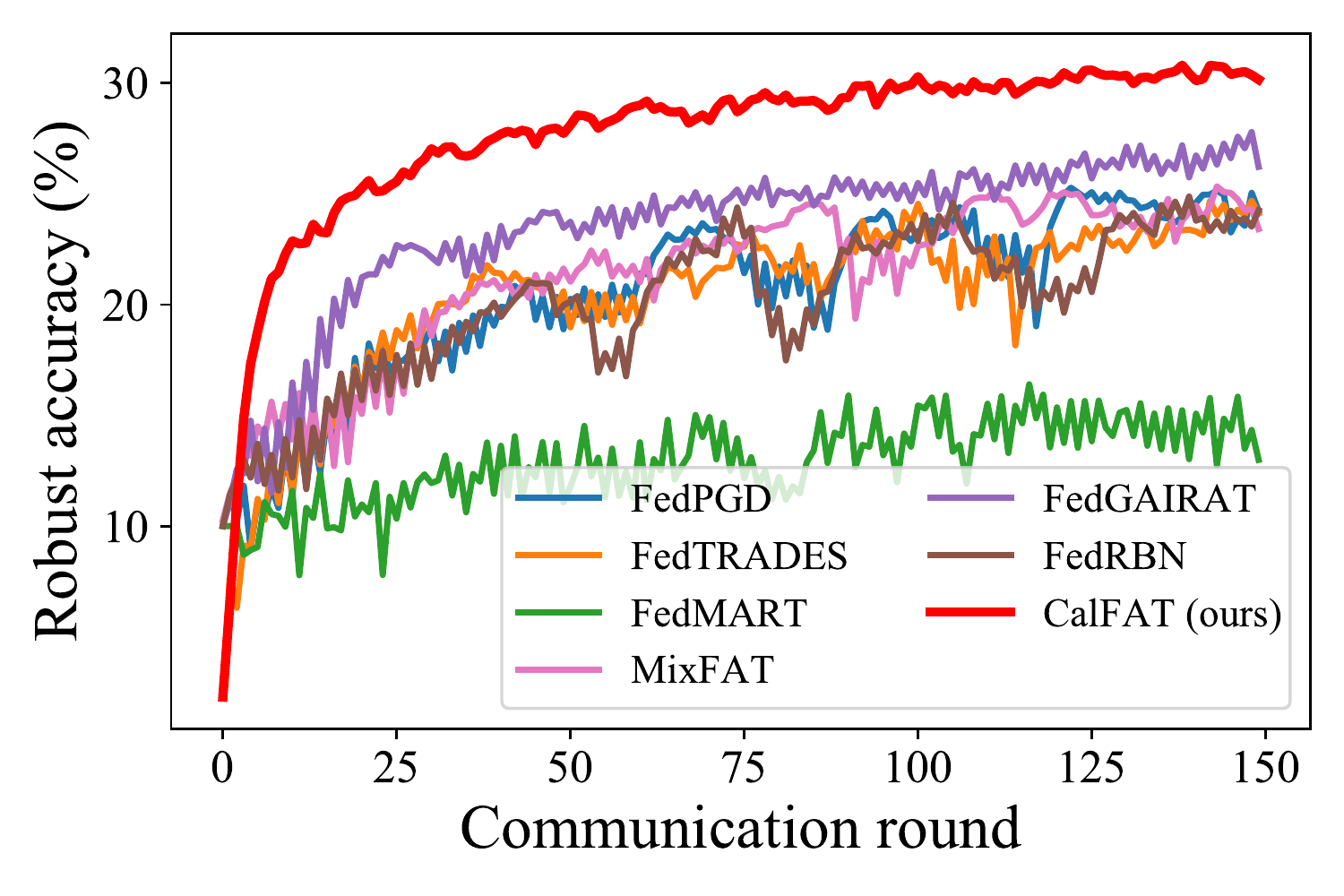}
\caption{Robust accuracy (against PGD-20 attack) of different methods on CIFAR10 dataset. 
}
\label{fig:learning_curve}
\end{minipage}
\hfill
\begin{minipage}{0.61\textwidth}
\centering
\includegraphics[height=2.9cm]{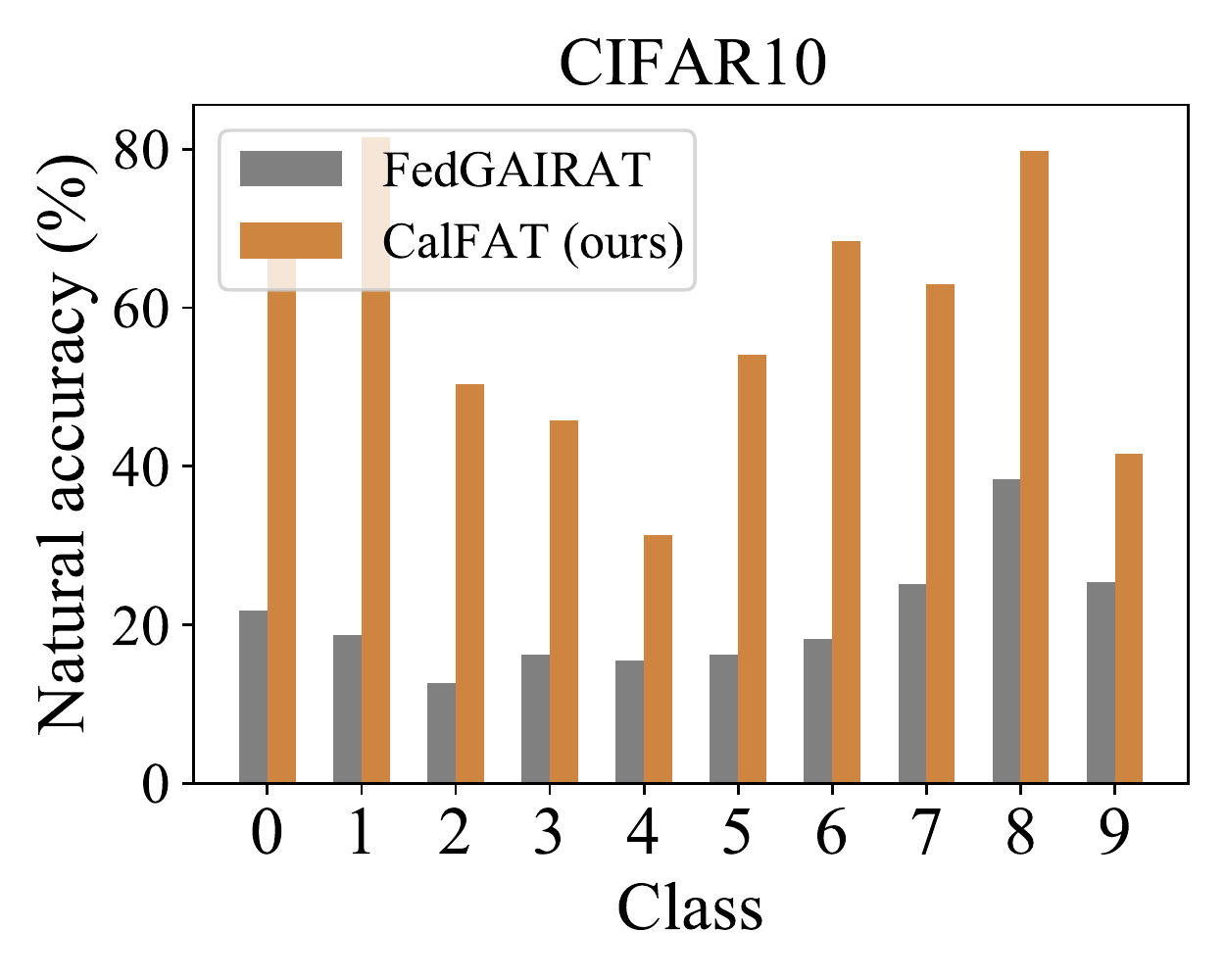}
\includegraphics[height=2.9cm]{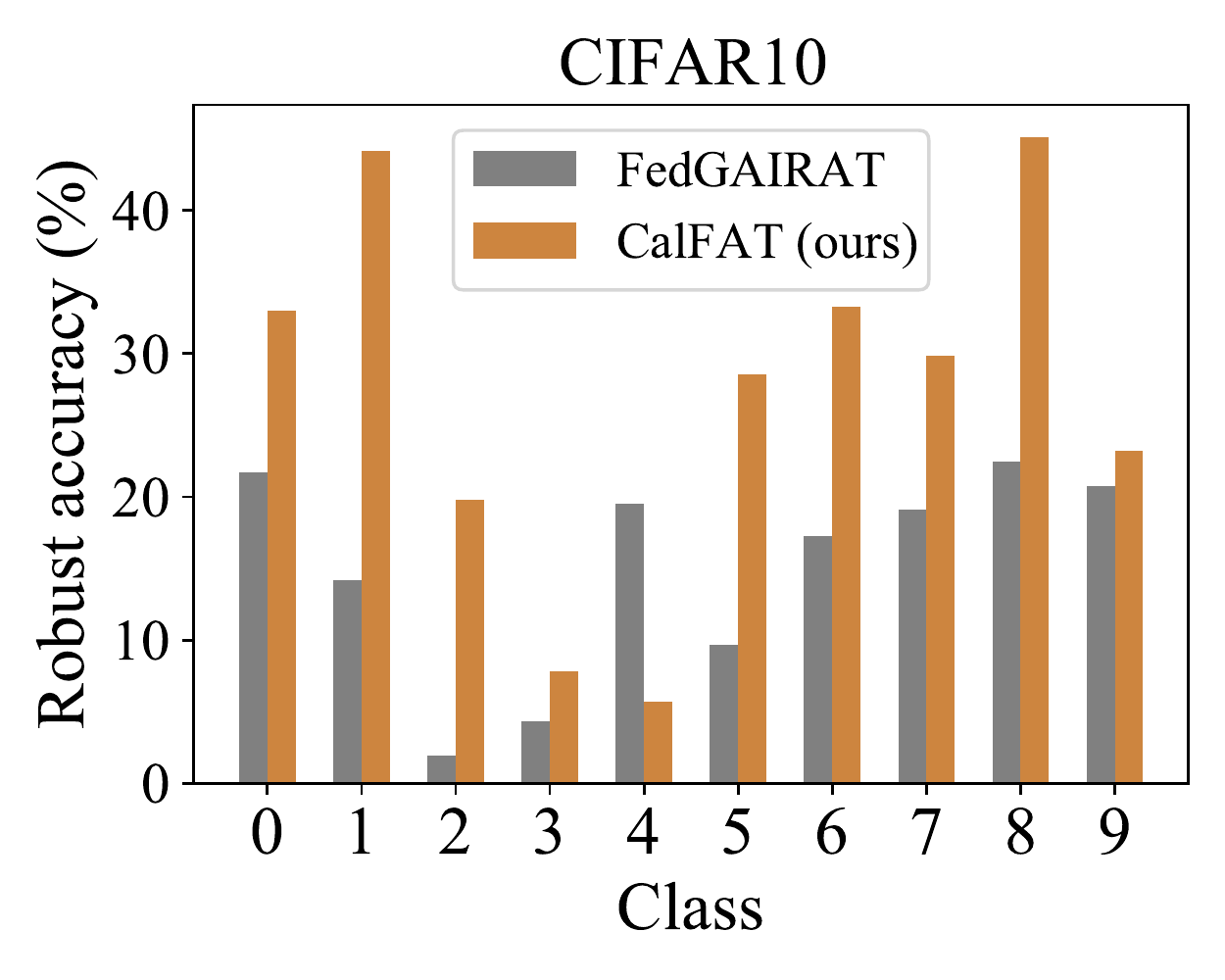}
\caption{Per-class natural accuracy and robust accuracy (against PGD-20 attack) of CalFAT and the best baseline (FedGAIRAT) on CIFAR10 dataset.
}
\label{fig:perclass_cifar10}
\end{minipage}
\end{figure}

\paragraph{Learning curves of different methods.}
To visually compare our CalFAT with all the baselines, we plot the learning curves (i.e., performance across different communication rounds) of all methods in Figure~\ref{fig:observation} and Figure~\ref{fig:learning_curve}. As can be observed, CalFAT achieves the best natural accuracy and robust accuracy across almost the entire training process, which indicates that the design of our CalFAT is profitable for different federated learning stages.
\begin{wraptable}{r}{8cm}
\caption{Combining FAT methods with different losses.
}
\label{tbl:longtail}
\centering
\scalebox{0.735}{
\begin{tabular}{c|cc}
\toprule
Metric & Natural & PGD-20 \\
\midrule
MixFAT & 53.35	$\pm$ 0.11 & 26.27	$\pm$ 0.11 \\
MixFAT + LogitAdj & 57.53	$\pm$ 0.21 & 27.65	$\pm$ 0.16 \\
MixFAT + RoBal & 58.25	$\pm$ 0.13 & 27.86	$\pm$ 0.10 \\
MixFAT + Calibration (ours) & \textbf{60.23}	$\pm$ 0.19 & \textbf{28.67}	$\pm$ 0.14 \\
\midrule
FedPGD & 46.96	$\pm$ 0.16 & 26.74	$\pm$ 0.18 \\
FedPGD + LogitAdj & 59.79	$\pm$ 0.15 & 28.84	$\pm$ 0.12 \\
FedPGD + RoBal & 61.48	$\pm$ 0.07 & 29.51	$\pm$ 0.07 \\
FedPGD + Calibration (ours) & \textbf{63.91}	$\pm$ 0.13 & \textbf{30.72}	$\pm$ 0.16 \\
\midrule
FedTRADES & 46.06	$\pm$ 0.12 & 26.31	$\pm$ 0.12 \\
FedTRADES + LogitAdj & 58.26	$\pm$ 0.20 & 27.92	$\pm$ 0.19 \\
FedTRADES + RoBal & 59.25	$\pm$ 0.23 & 28.63	$\pm$ 0.08 \\
FedTRADES + Calibration (ours) & \textbf{63.12}	$\pm$ 0.10 & \textbf{30.27}	$\pm$ 0.23 \\
\midrule
FedMART & 25.67	$\pm$ 0.21 & 18.10	$\pm$ 0.22 \\
FedMART + LogitAdj & 42.01	$\pm$ 0.10 & 24.92	$\pm$ 0.02 \\
FedMART + RoBal & 44.26	$\pm$ 0.22 & 25.57	$\pm$ 0.17 \\
FedMART + Calibration (ours) & \textbf{48.85}	$\pm$ 0.08 & \textbf{27.19}	$\pm$ 0.11 \\
\midrule
CalFAT (ours) & \textbf{64.69}	$\pm$ 0.08 & \textbf{31.12}	$\pm$ 0.11 \\
\bottomrule
\end{tabular}}
\vspace{-5pt}
\end{wraptable}
Moreover, our CalFAT is fairly stable during the whole training process while the accuracy curves of other baselines oscillate strongly. Such oscillations lead to bad convergence and low performance.
We hypothesize that the heterogeneity of local models in the baseline methods is the main cause of the unstable training.

\paragraph{Combining calibration loss with other FAT methods.}
To further show the effectiveness of our calibration loss, we combine it with four FAT methods (MixFAT, FedPGD, FedTRADES, and FedMART) and name them MixFAT + Calibration, FedPGD + Calibration, FedTRADES + Calibration, and FedMART + Calibration, respectively. We compare these calibration loss-based FAT methods with their original versions in Table~\ref{tbl:longtail}. It is evident that, by introducing our calibration loss into their objectives, all FAT methods can be improved. 
These results confirm the importance of class calibration for FAT with label skewness.

\paragraph{Comparison with state-of-the-art long-tail learning methods.}
We also compare our calibration loss with the losses used by long-tail learning methods LogitAdj \cite{menon2020long} and RoBal~\cite{wu2021adversarial}.
In particular, we combine four FAT methods (MixFAT, FedPGD, FedTRADES, and FedMART) with the above three losses (LogitAdj loss, RoBal loss, and our calibration loss), and train the models following the same default setting.
As shown in Table~\ref{tbl:longtail}, both LogitAdj-based methods and RoBal-based methods have lower natural and robust accuracies than calibration loss-based methods.
This indicates that our calibration loss is more suitable for FAT than other long-tail learning losses.

\subsection{Performance on Different Classes}
\label{sec:exp-diff-class}

We further compare the per-class performance of our CalFAT with the best baseline FedGAIRAT.
First, we use a well-trained model to initialize a global model. Second, the global model distributes the model parameter to all clients. Third, the local clients train their local models with their local data for 1 epoch. Then, we report the per-class average performance of all clients for each class. For fair comparison, we use the same well-trained model for initialization and the same data partition on each client for CalFAT and FedGAIRAT.

In Figure~\ref{fig:perclass_cifar10}, we report the per-class natural and robust accuracies of CalFAT and FedGAIRAT on CIFAR10. As shown in these figures, the average performance of most classes of CalFAT is much higher than FedGAIRAT.
We also report the per-class performance of each client on CIFAR10 in Appendix~\ref{appsec:per_class_different_client}. 
In FedGAIRAT, due to the highly skewed label distribution, the prediction of each client is highly biased to the majority classes, which leads to high performance on the majority classes and low performance (even 0\% accuracy) on the minority classes.
By contrast, in CalFAT, each client has higher performance on most classes. This verifies that the calibrated cross-entropy loss can indeed improve the performance on the minority classes, and further improve the overall performance of the model.
Moreover, we report the per-class average performance on SVHN in Appendix~\ref{appsec:per_class_avg}. Our CalFAT also outperforms the best baseline across most of the classes on SVHN.

\subsection{Results on Different FL Frameworks and Network Architectures}
\label{sec:exp-fl-network}
\paragraph{Evaluation on different FL frameworks.}
Besides FedAvg~\cite{mcmahan2017communication}, we also conduct experiments on other FL frameworks, i.e., FedProx~\cite{li2018federated} and Scaffold~\cite{karimireddy2020scaffold}. The results for all methods on FedProx and Scaffold are given in Table~\ref{tbl:framework}. It shows that our CalFAT exhibits better natural and robust accuracies than all baseline methods on all FL frameworks, which indicates the high comparability of our CalFAT with different aggregation algorithms.

\begin{table}[t]
\caption{Natural and robust accuracy (\%) across different FL frameworks on CIFAR10 dataset. 
}
\label{tbl:framework}
\centering
\resizebox{\textwidth}{!}{
\begin{tabular}{c|ccc|ccc}
\toprule
FL   framework &  \multicolumn{3}{c|}{FedProx} & \multicolumn{3}{c}{Scaffold} \\
\midrule
Metric & Natural & PGD-20 & AA & Natural & PGD-20 & AA \\
\midrule
MixFAT & 53.75	$\pm$ 0.16 & 29.61	$\pm$ 0.19 & 21.59	$\pm$ 0.27 & 55.27	$\pm$ 0.20 & 28.78	$\pm$ 0.15 & 21.26	$\pm$ 0.11 \\
FedPGD & 49.57	$\pm$ 0.18 & 28.48	$\pm$ 0.17 & 21.31	$\pm$ 0.18 & 49.52	$\pm$ 0.14 & 27.46	$\pm$ 0.21 & 20.27	$\pm$ 0.15 \\
FedTRADES & 48.14	$\pm$ 0.20 & 27.75	$\pm$ 0.17 & 21.13	$\pm$ 0.21 & 47.78	$\pm$ 0.23 & 27.31	$\pm$ 0.16 & 20.04	$\pm$ 0.16 \\
FedMART & 28.32	$\pm$ 0.22 & 19.32	$\pm$ 0.23 & 15.91	$\pm$ 0.25 & 27.80	$\pm$ 0.17 & 20.03	$\pm$ 0.26 & 16.85	$\pm$ 0.15 \\
FedGAIRAT & 49.61	$\pm$ 0.20 & 29.34	$\pm$ 0.11 & 21.33	$\pm$ 0.18 & 49.54	$\pm$ 0.21 & 27.23	$\pm$ 0.25 & 20.16	$\pm$ 0.09 \\
FedRBN & 47.26	$\pm$ 0.13 & 26.63	$\pm$ 0.15 & 20.46	$\pm$ 0.06 & 49.77	$\pm$ 0.09 & 28.37	$\pm$ 0.12 & 20.32	$\pm$ 0.06 \\
CalFAT & \textbf{66.32}	$\pm$ 0.08 & \textbf{32.79}	$\pm$ 0.13 & \textbf{22.83}	$\pm$ 0.11 & \textbf{67.16}	$\pm$ 0.06 & \textbf{32.94}	$\pm$ 0.06 & \textbf{21.94}	$\pm$ 0.05 \\
\bottomrule
\end{tabular}}
\end{table}

\paragraph{Evaluation on different network architectures.}
We also compare CalFAT with baselines on different network architectures, i.e., CNN~\cite{mcmahan2017communication}, VGG-8~\cite{simonyan2014very}, and ResNet-18~\cite{he2016deep}.
For CNN, we use the same architecture as~\cite{mcmahan2017communication}.
VGG-8 and ResNet-18 are two widely used architectures in deep learning.
The results on CIFAR10 dataset are shown in Appendix~\ref{appsec:arch}. 
CalFAT outperforms all baselines, which further 
validates the superiority of CalFAT with different network architectures. 

\subsection{Feature Visualization}
\label{sec:visualization}
To better understand the efficacy of CalFAT, we visualize the learned features extracted from the second last layer of FedTRADES (the best baseline) and CalFAT trained on SVHN dataset in Appendix~\ref{appsec:tsne}. 
The features are projected into a 2-dimensional space via t-SNE~\cite{van2008visualizing}. 
It shows that samples from different classes are mixed together in FedTRADES, indicating its low performance. 
For instance, Class 6 (pink) and Class 8 (khaki) are hard to separate in FedTRADES while these 2 classes can be well separated by CalFAT.
This illustration verifies that the server cannot learn a global model with good inter-class separability if the local models are heterogeneous. By contrast, CalFAT can well separate different classes thus can achieve better overall performance.

\subsection{Results under IID settings}
Besides the non-IID setting, we also conduct an experiment under the IID setting. The results are shown in Appendix \ref{appsec:iid} where it shows that our CalFAT achieves the best robustness (under PGD-20 attack).
Compared to the non-IID setting, all FAT methods demonstrate much better performance under the IID setting. This indicates that existing FAT methods can easily handle IID data yet face substantial challenges when the data is non-IID.

\subsection{Ablation Studies}
\label{sec:exp-ablation}

\paragraph{Impact of the number of clients.}
To show the generality of CalFAT, we train CalFAT with different numbers of clients $m$.
Table~\ref{apptbl:clients} in Appendix \ref{appsec:client_number} reports the results for $m=\{20,50,100\}$.
As expected, CalFAT achieves the best performance across all $m$. 
As $m$ increases, the performance of all methods decreases. We conjecture that this is because more clients in FAT makes the training harder to converge.
However, our CalFAT can still achieve 41.23\% natural accuracy when there are 100 clients, outperforming other baselines by a large margin.

\paragraph{Impact of skewed label distribution.}
We observe that the performance of FAT defense is closely related to label skewness. We thus investigate the impact of skewed label distribution by varying the Dirichlet parameter $\beta=\{0.05,0.2,0.3\}$ and report the results on CIFAR10 in Table~\ref{apptbl:beta} in Appendix \ref{appsec:skew_level}.
Not surprisingly, our CalFAT outperforms all baselines under all $\beta$'s. 
This further verifies the consistent effectiveness of CalFAT under different levels of label skewness.

Note that as $\beta$ decreases (i.e., the labels on each client are more imbalanced), the performance of all methods drop rapidly. For example, the natural accuracy of FedMART drops from 38.38\% to 29.84\% as $\beta$ decreases from 0.2 to 0.05. This indicates that all methods are hard to train a good model in extremely skewed label distribution scenarios. However, our CalFAT still achieves 61.00\% natural accuracy and 32.40\% robust accuracy (against FGSM attack) when $\beta=0.05$, which are much higher than all the baselines.

\paragraph{Contribution of the calibrated loss functions.}
As shown in Eq.~\eqref{eq:ce_ours} and Eq.~\eqref{eq:kl_ours},
for each client $i$, we have two new loss functions: a CCE loss $\ell_{cce}(\cdot,\cdot,\cdot)$ for optimization and a CKL loss $\ell_{ckl}(\cdot,\cdot,\cdot)$ for generating the adversarial examples. 
This naturally raises a question: how do these two loss functions contribute to CalFAT? 
To answer this question, we conduct leave-one-out tests by removing the CCE loss (w/o $\ell_{cce}(\cdot,\cdot,\cdot)$) or removing the CKL loss (w/o $\ell_{ckl}(\cdot,\cdot,\cdot)$) from the overall optimization objective.
As illustrated in Appendix \ref{appsec:loss_contribution}, w/o $\ell_{cce}(\cdot,\cdot,\cdot)$ leads to poor performance, which implies that CCE loss plays an important role in CalFAT.
Besides, if we only use the CCE loss (\ie w/o $\ell_{ckl}(\cdot,\cdot,\cdot)$), we can obtain a much better performance, but it still underperforms CalFAT. 
All these results indicate that the CCE loss is the most important part of CalFAT, whilst the CKL loss can further increase the performance of CalFAT.
The combination of both loss functions leads to the best performance.

\paragraph{Impact of the ratio of adversarial data.}
Here, we conduct experiments with different ratios of adversarial data used in CalFAT and report the  robust accuracy (against PGD-20 attack) in Appendix \ref{appsec:portion}.
Ratios $r$=0 and $r$=1 stand for training the model on only natural data and only adversarial data, respectively. Overall, $r$=1 produces the best robustness, meaning that training on only adversarial data can better enhance the adversarial robustness of our CalFAT.

\section{Conclusion} 
In this paper, we studied the challenging problem of Federated Adversarial Training (FAT) with label skewness and proposed a novel Calibrated Federated Adversarial Training (CalFAT) to simultaneously achieve stable training, better convergence, and natural accuracy and robustness in FL. 
CalFAT calibrates the model prediction and trains homogeneous local models across different clients by automatically assigning higher scores to the minority classes.
Extensive experiments on multiple datasets under various settings validate the effectiveness of CalFAT. Our work can serve as a simple but strong baseline for accurate and robust FAT.
For future work, we will continue to improve FAT under other non-IID settings such as feature skewness and quantity skewness \cite{hsieh2020skew, zhu2021noniidtype}.


\section*{Acknowledgement}
This work is funded by Sony AI.
This work is also supported by the National Key R\&D Program of China (Grant No. 2021ZD0112804) and the National Natural Science Foundation of China (Grant No. 62276067).

\bibliography{references}

\begin{thebibliography}{10}

\bibitem{bickel2015statistics}
Peter~J Bickel and Kjell~A Doksum.
\newblock {\em Mathematical statistics: basic ideas and selected topics,
  volumes I-II package}.
\newblock CRC Press, 2015.

\bibitem{carlini2017towards}
Nicholas Carlini and David Wagner.
\newblock Towards evaluating the robustness of neural networks.
\newblock In {\em 2017 ieee symposium on security and privacy (sp)}, pages
  39--57. IEEE, 2017.

\bibitem{carmon2019unlabeled}
Yair Carmon, Aditi Raghunathan, Ludwig Schmidt, Percy Liang, and John~C Duchi.
\newblock Unlabeled data improves adversarial robustness.
\newblock In {\em Proceedings of the 33rd International Conference on Neural
  Information Processing Systems}, pages 11192--11203, 2019.

\bibitem{chen2022decision}
Chen Chen, Jingfeng Zhang, Xilie Xu, Lingjuan Lyu, Chaochao Chen, Tianlei Hu,
  and Gang Chen.
\newblock Decision boundary-aware data augmentation for adversarial training.
\newblock {\em IEEE Transactions on Dependable and Secure Computing}, 2022.

\bibitem{croce2020reliable}
Francesco Croce and Matthias Hein.
\newblock Reliable evaluation of adversarial robustness with an ensemble of
  diverse parameter-free attacks.
\newblock In {\em International conference on machine learning}, pages
  2206--2216, 2020.

\bibitem{deng2009imagenet}
Jia Deng, Wei Dong, Richard Socher, Li-Jia Li, Kai Li, and Li~Fei-Fei.
\newblock Imagenet: A large-scale hierarchical image database.
\newblock In {\em 2009 IEEE conference on computer vision and pattern
  recognition}, pages 248--255, 2009.

\bibitem{guo2017interpret}
Chuan Guo, Geoff Pleiss, Yu~Sun, and Kilian~Q Weinberger.
\newblock On calibration of modern neural networks.
\newblock In {\em International Conference on Machine Learning}, pages
  1321--1330. PMLR, 2017.

\bibitem{he2016deep}
Kaiming He, Xiangyu Zhang, Shaoqing Ren, and Jian Sun.
\newblock Deep residual learning for image recognition.
\newblock In {\em Proceedings of the IEEE conference on computer vision and
  pattern recognition}, pages 770--778, 2016.

\bibitem{hong2021federated}
Junyuan Hong, Haotao Wang, Zhangyang Wang, and Jiayu Zhou.
\newblock Federated robustness propagation: Sharing adversarial robustness in
  federated learning.
\newblock {\em arXiv preprint arXiv:2106.10196}, 2021.

\bibitem{hsieh2020skew}
Kevin Hsieh, Amar Phanishayee, Onur Mutlu, and Phillip Gibbons.
\newblock The non-iid data quagmire of decentralized machine learning.
\newblock In {\em International Conference on Machine Learning}, pages
  4387--4398. PMLR, 2020.

\bibitem{karimireddy2020scaffold}
Sai~Praneeth Karimireddy, Satyen Kale, Mehryar Mohri, Sashank Reddi, Sebastian
  Stich, and Ananda~Theertha Suresh.
\newblock Scaffold: Stochastic controlled averaging for federated learning.
\newblock In {\em International Conference on Machine Learning}, pages
  5132--5143, 2020.

\bibitem{klenke2013bayes}
Achim Klenke.
\newblock {\em Probability theory: a comprehensive course}.
\newblock Springer Science \& Business Media, 2013.

\bibitem{krizhevsky2009learning}
Alex Krizhevsky, Geoffrey Hinton, et~al.
\newblock Learning multiple layers of features from tiny images.
\newblock In {\em Technical report}, 2009.

\bibitem{krizhevsky2012imagenet}
Alex Krizhevsky, Ilya Sutskever, and Geoffrey~E Hinton.
\newblock Imagenet classification with deep convolutional neural networks.
\newblock {\em Advances in neural information processing systems},
  25:1097--1105, 2012.

\bibitem{kurakin2016adversarial}
Alexey Kurakin, Ian Goodfellow, and Samy Bengio.
\newblock Adversarial machine learning at scale.
\newblock {\em arXiv preprint arXiv:1611.01236}, 2016.

\bibitem{li2018federated}
Tian Li, Anit~Kumar Sahu, Manzil Zaheer, Maziar Sanjabi, Ameet Talwalkar, and
  Virginia Smith.
\newblock Federated optimization in heterogeneous networks.
\newblock {\em arXiv preprint arXiv:1812.06127}, 2018.

\bibitem{li2020fedprox}
Tian Li, Anit~Kumar Sahu, Manzil Zaheer, Maziar Sanjabi, Ameet Talwalkar, and
  Virginia Smith.
\newblock Federated optimization in heterogeneous networks.
\newblock {\em Proceedings of Machine Learning and Systems}, 2:429--450, 2020.

\bibitem{li2021anti}
Yige Li, Xixiang Lyu, Nodens Koren, Lingjuan Lyu, Bo~Li, and Xingjun Ma.
\newblock Anti-backdoor learning: Training clean models on poisoned data.
\newblock {\em Advances in Neural Information Processing Systems}, 34, 2021.

\bibitem{luo2019skeweg}
Jiahuan Luo, Xueyang Wu, Yun Luo, Anbu Huang, Yunfeng Huang, Yang Liu, and
  Qiang Yang.
\newblock Real-world image datasets for federated learning.
\newblock {\em arXiv preprint arXiv:1910.11089}, 2019.

\bibitem{lyu2020privacy}
Lingjuan Lyu, Han Yu, Xingjun Ma, Chen Chen, Lichao Sun, Jun Zhao, Qiang Yang,
  and Philip~S Yu.
\newblock Privacy and robustness in federated learning: Attacks and defenses.
\newblock {\em arXiv preprint arXiv:2012.06337}, 2020.

\bibitem{lyu2020threats}
Lingjuan Lyu, Han Yu, Jun Zhao, and Qiang Yang.
\newblock Threats to federated learning.
\newblock In {\em Federated Learning}, pages 3--16. Springer, 2020.

\bibitem{madry2017towards}
Aleksander Madry, Aleksandar Makelov, Ludwig Schmidt, Dimitris Tsipras, and
  Adrian Vladu.
\newblock Towards deep learning models resistant to adversarial attacks.
\newblock {\em arXiv preprint arXiv:1706.06083}, 2017.

\bibitem{mcmahan2017communication}
Brendan McMahan, Eider Moore, Daniel Ramage, Seth Hampson, and Blaise~Aguera
  y~Arcas.
\newblock Communication-efficient learning of deep networks from decentralized
  data.
\newblock In {\em Artificial Intelligence and Statistics}, pages 1273--1282.
  PMLR, 2017.

\bibitem{menon2020long}
Aditya~Krishna Menon, Sadeep Jayasumana, Ankit~Singh Rawat, Himanshu Jain,
  Andreas Veit, and Sanjiv Kumar.
\newblock Long-tail learning via logit adjustment.
\newblock {\em arXiv preprint arXiv:2007.07314}, 2020.

\bibitem{netzer2011reading_SVHN}
Yuval Netzer, Tao Wang, Adam Coates, Alessandro Bissacco, Bo~Wu, and Andrew~Y
  Ng.
\newblock Reading digits in natural images with unsupervised feature learning.
\newblock In {\em NeurIPS Workshop on Deep Learning and Unsupervised Feature
  Learning}, 2011.

\bibitem{rice2020overfitting}
Leslie Rice, Eric Wong, and J~Zico Kolter.
\newblock Overfitting in adversarially robust deep learning.
\newblock In {\em ICML}, 2020.

\bibitem{simonyan2014very}
Karen Simonyan and Andrew Zisserman.
\newblock Very deep convolutional networks for large-scale image recognition.
\newblock {\em arXiv preprint arXiv:1409.1556}, 2014.

\bibitem{tan2022fedproto}
Yue Tan, Guodong Long, Lu~Liu, Tianyi Zhou, Qinghua Lu, Jing Jiang, and Chengqi
  Zhang.
\newblock Fedproto: Federated prototype learning across heterogeneous clients.
\newblock In {\em AAAI Conference on Artificial Intelligence}, volume~36, pages
  8432--8440, 2022.

\bibitem{tan2022federated}
Yue Tan, Guodong Long, Jie Ma, Lu~Liu, Tianyi Zhou, and Jing Jiang.
\newblock Federated learning from pre-trained models: A contrastive learning
  approach.
\newblock In {\em First Workshop on Pre-training: Perspectives, Pitfalls, and
  Paths Forward at ICML 2022}.

\bibitem{tsipras2018robustness}
Dimitris Tsipras, Shibani Santurkar, Logan Engstrom, Alexander Turner, and
  Aleksander Madry.
\newblock Robustness may be at odds with accuracy.
\newblock In {\em International Conference on Learning Representations}, 2018.

\bibitem{van2008visualizing}
Laurens Van~der Maaten and Geoffrey Hinton.
\newblock Visualizing data using t-sne.
\newblock {\em Journal of machine learning research}, 9(11), 2008.

\bibitem{wald1949consistency}
Abraham Wald.
\newblock Note on the consistency of the maximum likelihood estimate.
\newblock {\em The Annals of Mathematical Statistics}, 20(4):595--601, 1949.

\bibitem{wang2021imbalanced}
Wentao Wang, Han Xu, Xiaorui Liu, Yaxin Li, Bhavani Thuraisingham, and Jiliang
  Tang.
\newblock Imbalanced adversarial training with reweighting.
\newblock {\em arXiv preprint arXiv:2107.13639}, 2021.

\bibitem{wang2020improving_MART}
Yisen Wang, Difan Zou, Jinfeng Yi, James Bailey, Xingjun Ma, and Quanquan Gu.
\newblock Improving adversarial robustness requires revisiting misclassified
  examples.
\newblock In {\em ICLR}, 2020.

\bibitem{wong2020fast_zico_kolter}
Eric Wong, Leslie Rice, and J.~Zico Kolter.
\newblock Fast is better than free: Revisiting adversarial training.
\newblock In {\em ICLR}, 2020.

\bibitem{wu2021adversarial}
Tong Wu, Ziwei Liu, Qingqiu Huang, Yu~Wang, and Dahua Lin.
\newblock Adversarial robustness under long-tailed distribution.
\newblock In {\em Proceedings of the IEEE/CVF Conference on Computer Vision and
  Pattern Recognition}, pages 8659--8668, 2021.

\bibitem{xu2021robust}
Han Xu, Xiaorui Liu, Yaxin Li, Anil Jain, and Jiliang Tang.
\newblock To be robust or to be fair: Towards fairness in adversarial training.
\newblock In {\em International Conference on Machine Learning}, pages
  11492--11501, 2021.

\bibitem{yang2019federated}
Qiang Yang, Yang Liu, Tianjian Chen, and Yongxin Tong.
\newblock Federated machine learning: Concept and applications.
\newblock {\em ACM Transactions on Intelligent Systems and Technology (TIST)},
  10(2):1--19, 2019.

\bibitem{yurochkin2019bayesian}
Mikhail Yurochkin, Mayank Agarwal, Soumya Ghosh, Kristjan Greenewald, Nghia
  Hoang, and Yasaman Khazaeni.
\newblock Bayesian nonparametric federated learning of neural networks.
\newblock In {\em International Conference on Machine Learning}, pages
  7252--7261, 2019.

\bibitem{zhang2019theoretically}
Hongyang Zhang, Yaodong Yu, Jiantao Jiao, Eric Xing, Laurent El~Ghaoui, and
  Michael Jordan.
\newblock Theoretically principled trade-off between robustness and accuracy.
\newblock In {\em International Conference on Machine Learning}, pages
  7472--7482. PMLR, 2019.

\bibitem{zhang2022qekd}
Jie Zhang, Chen Chen, Jiahua Dong, Ruoxi Jia, and Lingjuan Lyu.
\newblock Qekd: Query-efficient and data-free knowledge distillation from
  black-box models.
\newblock {\em arXiv preprint arXiv:2205.11158}, 2022.

\bibitem{zhang2022towards}
Jie Zhang, Bo~Li, Jianghe Xu, Shuang Wu, Shouhong Ding, Lei Zhang, and Chao Wu.
\newblock Towards efficient data free black-box adversarial attack.
\newblock In {\em Proceedings of the IEEE/CVF Conference on Computer Vision and
  Pattern Recognition}, pages 15115--15125, 2022.

\bibitem{zhang2022federated}
Jie Zhang, Zhiqi Li, Bo~Li, Jianghe Xu, Shuang Wu, Shouhong Ding, and Chao Wu.
\newblock Federated learning with label distribution skew via logits
  calibration.
\newblock In {\em International Conference on Machine Learning}, pages
  26311--26329, 2022.

\bibitem{zhang2021geometry}
Jingfeng Zhang, Jianing Zhu, Gang Niu, Bo~Han, Masashi Sugiyama, and Mohan
  Kankanhalli.
\newblock Geometry-aware instance-reweighted adversarial training.
\newblock In {\em ICLR}, 2021.

\bibitem{zhou2021adversarially}
Yao Zhou, Jun Wu, and Jingrui He.
\newblock Adversarially robust federated learning for neural networks, 2021.

\bibitem{zhu2021noniidtype}
Hangyu Zhu, Jinjin Xu, Shiqing Liu, and Yaochu Jin.
\newblock Federated learning on non-iid data: A survey.
\newblock {\em Neurocomputing}, 465:371--390, 2021.

\bibitem{zizzo2020fat}
Giulio Zizzo, Ambrish Rawat, Mathieu Sinn, and Beat Buesser.
\newblock Fat: Federated adversarial training.
\newblock {\em arXiv preprint arXiv:2012.01791}, 2020.

\end{thebibliography}
\bibliographystyle{plain}






\section*{Checklist}


\begin{enumerate}

\item For all authors...
\begin{enumerate}
  \item Do the main claims made in the abstract and introduction accurately reflect the paper's contributions and scope?
    \answerYes{}
  \item Did you describe the limitations of your work?
    \answerYes{}
  \item Did you discuss any potential negative societal impacts of your work?
    \answerNo{}
  \item Have you read the ethics review guidelines and ensured that your paper conforms to them?
    \answerYes{}
\end{enumerate}

\item If you are including theoretical results...
\begin{enumerate}
  \item Did you state the full set of assumptions of all theoretical results?
    \answerYes{}
        \item Did you include complete proofs of all theoretical results?
    \answerYes{}
\end{enumerate}

\item If you ran experiments...
\begin{enumerate}
  \item Did you include the code, data, and instructions needed to reproduce the main experimental results (either in the supplemental material or as a URL)?
    \answerYes{}
  \item Did you specify all the training details (e.g., data splits, hyperparameters, how they were chosen)?
    \answerYes{}
        \item Did you report error bars (e.g., with respect to the random seed after running experiments multiple times)?
    \answerNo{}
        \item Did you include the total amount of compute and the type of resources used (e.g., type of GPUs, internal cluster, or cloud provider)?
    \answerYes{}
\end{enumerate}

\item If you are using existing assets (e.g., code, data, models) or curating/releasing new assets...
\begin{enumerate}
  \item If your work uses existing assets, did you cite the creators?
    \answerYes{}
  \item Did you mention the license of the assets?
    \answerNA{}{}
  \item Did you include any new assets either in the supplemental material or as a URL?
    \answerNo{}
  \item Did you discuss whether and how consent was obtained from people whose data you're using/curating?
    \answerNA{}
  \item Did you discuss whether the data you are using/curating contains personally identifiable information or offensive content?
    \answerNA{}
\end{enumerate}

\item If you used crowdsourcing or conducted research with human subjects...
\begin{enumerate}
  \item Did you include the full text of instructions given to participants and screenshots, if applicable?
    \answerNA{}
  \item Did you describe any potential participant risks, with links to Institutional Review Board (IRB) approvals, if applicable?
    \answerNA{}
  \item Did you include the estimated hourly wage paid to participants and the total amount spent on participant compensation?
    \answerNA{}
\end{enumerate}

\end{enumerate}
\clearpage


\appendix


\section{Proof of Lemma 1}
\label{appsec:proof_lemma}
\textbf{Lemma 1} (Non-identical class probabilities). 
\textit{If the label distribution across the clients is skewed and the class conditionals have the same support, then the class probabilities $\{p_i(y\mid x)\mid i\in[m]\}$ are non-identical, \ie{}
for all $i\ne u$ and $i,u\in[m]$, there exists $x$, $y$ such that $p_i(y\mid x)\ne p_u(y\mid x)$.
}

\begin{proof}
\emph{Case 1:} For all $y\in[C]$, $p_i(y),p_u(y)>0$  or $p_i(y)=p_u(y)=0$ .

We prove the result by contradiction.
Assume that $p_i(y\mid x)= p_u(y\mid x)$ holds for all $x$, $y\in[C]$.

Consider $y\in[C]$ so that $p_i(y),p_u(y)>0$.
For all $x$, $p_i(x\mid y)>0$
\begin{align}
    p_i(y\mid x)=p_u(y\mid x).
\end{align}
According to the Bayes' rule,
\begin{align}
    \frac{p_i(x\mid y)p_i(y)}{p_i(x)}=\frac{p_u(x\mid y)p_u(y)}{p_u(x)}
\end{align}
Cancel the $p_i(x\mid y)=p_u(x\mid y)\ne 0$ and obtain
\begin{align}
    \frac{p_i(y)}{p_i(x)}=\frac{p_u(y)}{p_u(x)}.
\end{align}
Take the reciprocal of both sides,
\begin{align}
    \frac{p_i(x)}{p_i(y)}=\frac{p_u(x)}{p_u(y)}.
\end{align}
Calculate the integral of both sides:
\begin{align}
    \int\frac{p_i(x)}{p_i(y)}dx{}&=\int\frac{p_u(x)}{p_u(y)}dx,\\
    \Rightarrow\frac{1}{p_i(y)}{}&=\frac{1}{p_u(y)}\\
    \Rightarrow p_i(y){}&=p_u(y).
\end{align}
This result contradicts the fact that there exists $y\in[C]$ such that $p_i(y)=p_u(y)$.
Therefore, we conclude that the assumption must be false and that its opposite there exists $x$, $y\in[C]$ such that $p_i(y\mid x)\ne p_u(y\mid x)$ must be true in this case.

\emph{Case 2:} There exists $y\in[C]$ that satisfies $p_i(y)>0$, $p_u(y)=0$ or $p_i(y)=0$, $p_u(y)>0$.
Without loss of generality, we consider $p_i(y)>0$ and $p_u(y)=0$.

Take $x$ so that $p_i(x\mid y)>0$, then according to Bayes' formula,
\begin{align}
    p_i(y\mid x){}&=\frac{p_i(x\mid y)p_i(y)}{p_i(x)}> 0,\\
    p_u(y\mid x){}&=\frac{p_u(x\mid y)p_u(y)}{p_u(x)}=0.
\end{align}
Therefore, $p_i(y\mid x)\ne p_i(y\mid x)$, which completes the proof.

\end{proof}

\section{Proof of Proposition~\ref{prop:hetero}}
\label{appsec:proof_pro1}

\textbf{Proposition 1} (Heterogeneous local models). 
\textit{Assume the label distribution across the clients is skewed.
Let $\theta_i$ be the maximum likelihood estimate of $\theta_i^*$ in Eq. \eqref{eq:parameterize_naivefat} given local data at client $i$.
Then $s^2$ converges almost surely to a nonzero constant:
$$
    s^2\xrightarrow[]{\text{a.s.}} (s^*)^2\ne 0,
$$
where $\xrightarrow[]{\text{a.s.}}$ represents the almost sure convergence.}

\begin{proof}
According to the definition of sample variance, the convergence of local model parameters implies the convergence of $s^2$:
\begin{align}
    \left\{\lim_{n_1,\ldots,n_m\to\infty}s^2=(s^*)^2\right\}
    \supseteq
    \left\{\lim_{n_i\to\infty}\theta_i=\theta_i^*,\forall i\in[m]\right\}.
\end{align}
Then since probability is monotonic, we have
\begin{align}
\label{eq:app_mono}
    \Pr\left\{\lim_{n_1,\ldots,n_m\to\infty}s^2=(s^*)^2\right\}
    \ge\Pr\left\{\lim_{n_i\to\infty}\theta_i=\theta_i^*,\forall i\in[m]\right\}.
\end{align}
Since the sampling on different clients is independent, $\theta_i$ are independent, we have:
\begin{align}
\label{eq:app_indep}
    \Pr\left\{\lim_{n_i\to\infty}\theta_i=\theta_i^*,\forall i\in[m]\right\}
    =\prod_{i=1}^m\Pr\left\{\lim_{n_i\to\infty}\theta_i=\theta_i^*\right\}.
\end{align}
According to \cite{wald1949consistency}, the MLE $\theta_i$ is a consistent estimate of $\theta_i^*$:
\begin{align}
\label{eq:app_cons}
    \Pr\left\{\lim_{n_i\to\infty}\theta_i=\theta_i^*\right\}=1,\quad i\in[m].
\end{align}
By combining Eq. \eqref{eq:app_mono}, Eq. \eqref{eq:app_indep} and Eq. \eqref{eq:app_cons}, it follows that
\begin{align}
    \Pr\left\{\lim_{n_1,\ldots,n_m\to\infty}s^2=(s^*)^2\right\}\ge1,
\end{align}
which implies
\begin{align}
    \Pr\left\{\lim_{n_1,\ldots,n_m\to\infty}s^2=(s^*)^2\right\}=1
    \quad\Rightarrow\quad
    s^2\xrightarrow[]{\text{a.s.}}(s^*)^2.
\end{align}
\end{proof}

\section{Proof of Proposition \ref{prop:homo}}
\label{appsec:proof_pro2}

\textbf{Proposition 2} (Homogeneous local models). 
\textit{Assume the label distribution across the clients is skewed.
Let $\theta_i$ be the maximum likelihood estimate of $\theta^*$ in Eq. \eqref{eq:repara} given local data at client $i$. 
Then $s^2$ converges almost surely to zero:
$$
    s^2\xrightarrow[]{\text{a.s.}} 0.
$$
}


\begin{proof}
According to the definition of sample variance, the convergence of local model parameters implies the convergence of $s^2$:
\begin{align}
    \left\{\lim_{n_1,\ldots,n_m\to\infty}s^2=0\right\}
    \supseteq
    \left\{\lim_{n_i\to\infty}\theta_i=\theta^*,\forall i\in[m]\right\}.
\end{align}
Then since probability is monotonic, we have
\begin{align}
\label{eq:app_homo_mono}
    \Pr\left\{\lim_{n_1,\ldots,n_m\to\infty}s^2=0\right\}
    \ge\Pr\left\{\lim_{n_i\to\infty}\theta_i=\theta^*,\forall i\in[m]\right\}.
\end{align}
Since the sampling on different clients is independent, $\theta_i$ are independent, we have:
\begin{align}
\label{eq:app_homo_indep}
    \Pr\left\{\lim_{n_i\to\infty}\theta_i=\theta^*,\forall i\in[m]\right\}
    =\prod_{i=1}^m\Pr\left\{\lim_{n_i\to\infty}\theta_i=\theta^*\right\}.
\end{align}
According to \cite{wald1949consistency}, the MLE $\theta_i$ is a consistent estimate of $\theta^*$:
\begin{align}
\label{eq:app_homo_cons}
    \Pr\left\{\lim_{n_i\to\infty}\theta_i=\theta^*\right\}=1,\quad i\in[m].
\end{align}
By combining Eq. \eqref{eq:app_homo_mono}, Eq. \eqref{eq:app_homo_indep} and Eq. \eqref{eq:app_homo_cons}, it follows that
\begin{align}
    \Pr\left\{\lim_{n_1,\ldots,n_m\to\infty}s^2=0\right\}\ge1,
\end{align}
which implies
\begin{align}
    \Pr\left\{\lim_{n_1,\ldots,n_m\to\infty}s^2=0\right\}=1
    \quad\Rightarrow\quad
    s^2\xrightarrow[]{\text{a.s.}}0.
\end{align}
\end{proof}

\section{Experimental Setup and Additional Experiments}
\label{appsec:exp}

\subsection{Detailed Experimental Setup}
\label{appsec:exp_setting}

\paragraph{Datasets.}
Our experiments are conducted on 4 real-world datasets: CIFAR10~\cite{krizhevsky2009learning}, CIFAR100~\cite{krizhevsky2009learning}, SVHN~\cite{netzer2011reading_SVHN}, and ImageNet subset~\cite{deng2009imagenet}.
The ImageNet subset is generated according to~\cite{li2021anti}, which consists of 12 labels. We resize the original image (with size 224*224*3) to 64*64*3 for fast training. 

\paragraph{Data partition.}
\begin{wrapfigure}{h}{.5\textwidth}
    \vspace{-0.6cm}
	\centering
	\includegraphics[width=0.88\linewidth]{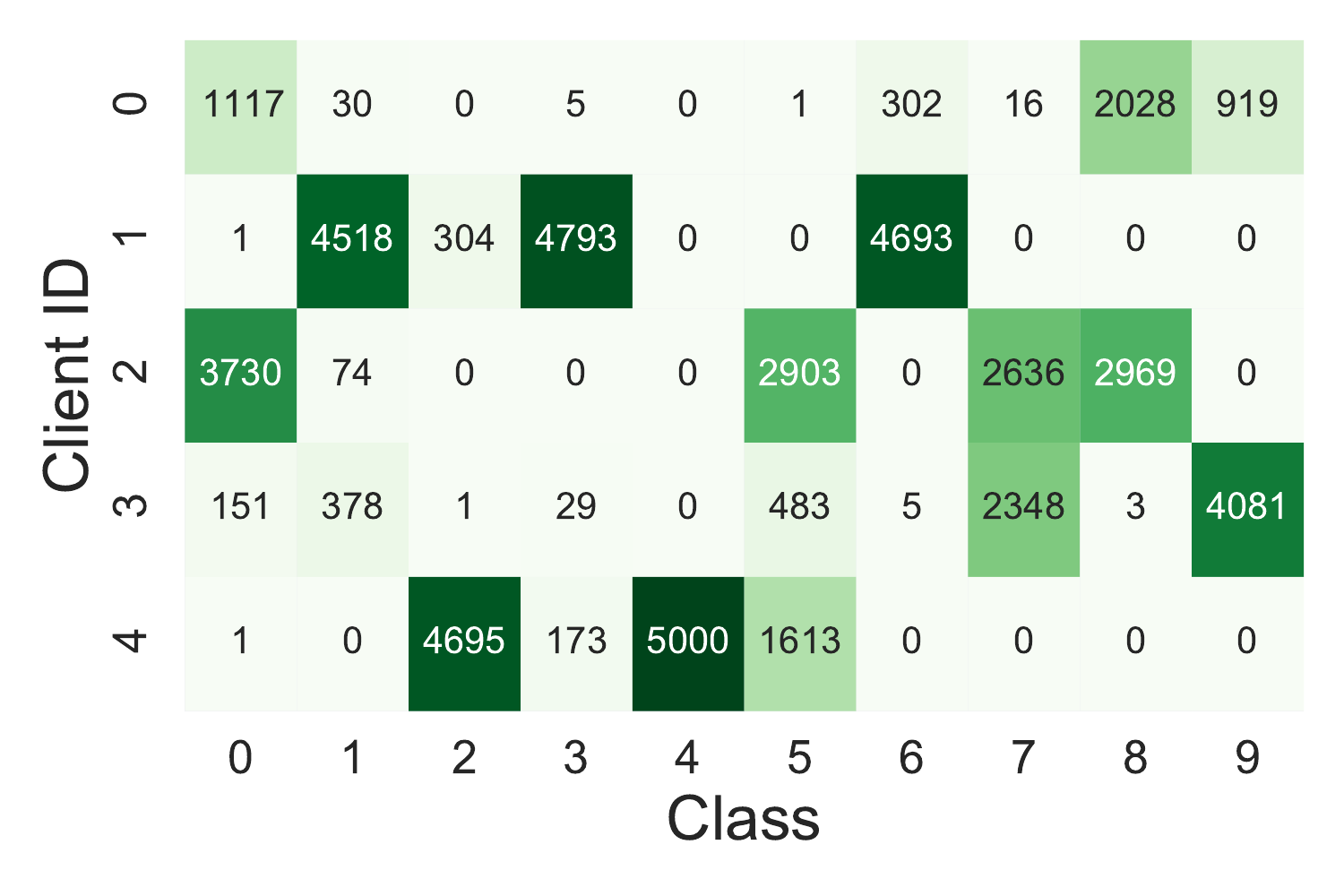}
	\caption{Label distribution of 
	CIFAR10 among 5 different clients.}
	\label{fig:data_partition}
\end{wrapfigure}
To simulate real-world statistical heterogeneity, we use Dirichlet distribution to generate non-IID data across clients~\cite{yurochkin2019bayesian}. 
In particular, we sample $p_i^{l} \sim Dir(\beta)$ and allocate a $p^l_i$ proportion of the data of label $l$ to client $i$, where $Dir(\beta)$ is the Dirichlet distribution with a concentration parameter $\beta$.
To simulate a highly skewed label distribution that widely exists in reality, we set $\beta=0.1$ as default. We visualize the label distribution of 5 clients on CIFAR10 dataset (when $\beta=0.1$) in Figure~\ref{fig:data_partition}. 
The number in the figure stands for the number of training samples associated with the corresponding label in one particular client. 
As shown in the figure, the label distribution is highly skewed and each client has relatively few data (even no data) on some classes.


\paragraph{Metric.}
For evaluation, we report the natural test accuracy (Natural) on natural test data and the robust test accuracy on adversarial test data. The adversarial test data are generated by FGSM (fast gradient sign method)~\cite{wong2020fast_zico_kolter}, BIM (basic iterative method with 20 steps)~\cite{kurakin2016adversarial}, PGD-20 (projected gradient descent with 20 steps)~\cite{madry2017towards}, CW (CW with 20 steps)~\cite{carlini2017towards}, and AA (auto attack)~\cite{croce2020reliable} with the same perturbation bound $\epsilon=8/255$. The step sizes for BIM, PGD-20 attack, and CW attack are $2/255$.

\paragraph{Setting.}
In our experiments, we consider $||\widetilde{x}-x||_{\infty}<\epsilon$ with the same $\epsilon$ for both training and evaluation.
To generate the most adversarial data to update the model, we follow the same setting as~\cite{rice2020overfitting}, i.e., we set the perturbation bound to $\epsilon=8/255$; PGD step number to $K=10$; and PGD step size to $\alpha=2/255$. 
We train the model by using SGD with momentum$=0.9$ and learning rate $\eta=0.01$.
The number of communication rounds is set to $T=150$ and the number of local epochs is set to $E=1$. 
All methods use FedAvg for aggregation and use the same CNN network~\cite{mcmahan2017communication} on CIFAR10, CIFAR100, and SVHN datasets.
We adopt Alexnet~\cite{krizhevsky2012imagenet} to train the ImageNet subset for all methods.
Recall that, compared with the cross-device setting, FAT matters more in the cross-silo setting, in which the number of clients is relatively \emph{small}, and each client has powerful computation resources to handle the computation cost of AT~\cite{lyu2020threats}. 
Thus, we set the number of clients to $m=5$ by default, and in each epoch, all clients are involved in the training. Experiments with more clients can be referred to Table~\ref{apptbl:clients} in Appendix \ref{appsec:client_number}.
The experiments are run on a server with Intel(R) Xeon(R) Gold 5218R CPU, 64GB RAM, and 8 Tesla V100 GPUs.

\subsection{Per-class Performance of Different Clients}
\label{appsec:per_class_different_client}
Table~\ref{apptbl:per_class} shows the per-class performance of different clients on CIFAR10 dataset. 
In FedGAIRAT, due to highly skewed label distribution, the prediction of each client is highly biased to the majority classes, leading to high performance on the majority classes and low performance (even 0\% accuracy) on the minority classes.
By contrast, in CalFAT, each client has higher performance on most classes. 
For example, on client 1, the accuracy of class 8 (96.56\%) of FedGAIRAT is higher than CalFAT, due to that the prediction is highly biased to class 8 on client 1 for FedGAIRAT. By contrast, the accuracy of other (minority) classes on client 1 of FedGAIRAT is much lower than CalFAT.
These results show that the calibrated cross-entropy loss can indeed improve the performance on minority classes, and further improve the overall performance of the model.

\begin{table}[h!]
\caption{Per-class natural accuracy and robust accuracy (against PGD-20 attack) of different clients on CIFAR10 dataset. }
\label{apptbl:per_class}
\centering
\resizebox{\textwidth}{!}{
\begin{tabular}{c|c|c|cccccccccc|c}
\toprule
\multicolumn{3}{c|}{Class} & 0 & 1 & 2 & 3 & 4 & 5 & 6 & 7 & 8 & 9 & Average \\
\midrule
\multirow{10}{*}{Natural} & \multirow{2}{*}{client 1} & FedGAIRAT & 47.45 & 0.00 & 0.00 & 0.00 & 0.00 & 0.00 & 16.91 & 0.00 & 96.56 & 27.30 & 18.82 \\
 &  & CalFAT(ours) & 56.61 & 88.06 & 54.41 & 27.66 & 37.67 & 65.16 & 52.02 & 74.03 & 89.02 & 54.09 & 59.87 \\
 & \multirow{2}{*}{client 2} & FedGAIRAT & 0.00 & 93.24 & 0.00 & 80.97 & 0.00 & 0.00 & 74.49 & 0.00 & 0.00 & 0.00 & 24.87 \\
 &  & CalFAT(ours) & 71.55 & 86.34 & 58.46 & 62.88 & 16.75 & 29.28 & 78.75 & 47.76 & 75.44 & 42.22 & 56.94 \\
 & \multirow{2}{*}{client 3} & FedGAIRAT & 57.41 & 0.00 & 0.00 & 0.00 & 0.00 & 69.81 & 0.07 & 69.96 & 95.01 & 0.00 & 29.23 \\
 &  & CalFAT(ours) & 90.11 & 82.41 & 41.48 & 40.41 & 17.42 & 64.65 & 74.96 & 58.18 & 74.25 & 22.08 & 56.60 \\
 & \multirow{2}{*}{client 4} & FedGAIRAT & 4.23 & 0.61 & 0.00 & 0.00 & 0.06 & 2.54 & 0.00 & 56.13 & 0.00 & 99.72 & 16.33 \\
 &  & CalFAT(ours) & 59.76 & 76.50 & 46.15 & 28.80 & 28.23 & 63.09 & 79.07 & 77.62 & 79.06 & 46.88 & 58.52 \\
 & \multirow{2}{*}{client 5} & FedGAIRAT & 0.00 & 0.00 & 63.33 & 0.00 & 77.49 & 8.86 & 0.00 & 0.00 & 0.00 & 0.00 & 14.97 \\
 &  & CalFAT(ours) & 70.48 & 74.60 & 51.58 & 69.32 & 56.94 & 48.69 & 57.86 & 57.65 & 80.16 & 43.06 & 61.03 \\
 \midrule
\multirow{10}{*}{Robust} & \multirow{2}{*}{client 1} & FedGAIRAT & 35.53 & 0.00 & 0.00 & 0.00 & 0.00 & 0.00 & 3.61 & 0.00 & 87.08 & 10.25 & 13.65 \\
 &  & CalFAT(ours) & 29.28 & 59.57 & 19.24 & 4.80 & 6.98 & 31.42 & 10.79 & 37.27 & 63.71 & 18.02 & 28.11 \\
 & \multirow{2}{*}{client 2} & FedGAIRAT & 0.00 & 71.03 & 0.06 & 21.54 & 0.04 & 0.00 & 82.64 & 0.00 & 0.00 & 0.00 & 17.53 \\
 &  & CalFAT(ours) & 35.61 & 72.05 & 15.45 & 22.06 & 5.24 & 7.89 & 55.68 & 23.54 & 39.47 & 6.17 & 28.32 \\
 & \multirow{2}{*}{client 3} & FedGAIRAT & 66.94 & 0.00 & 0.00 & 0.00 & 0.00 & 39.56 & 0.00 & 53.36 & 25.35 & 0.00 & 18.52 \\
 &  & CalFAT(ours) & 38.59 & 36.92 & 27.43 & 7.06 & 2.01 & 26.73 & 27.29 & 39.53 & 40.30 & 20.25 & 26.61 \\
 & \multirow{2}{*}{client 4} & FedGAIRAT & 6.02 & 0.00 & 0.00 & 0.00 & 0.00 & 0.12 & 0.00 & 42.34 & 0.00 & 93.73 & 14.22 \\
 &  & CalFAT(ours) & 17.37 & 11.05 & 9.17 & 1.17 & 3.35 & 47.13 & 55.97 & 27.78 & 56.53 & 51.64 & 28.12 \\
 & \multirow{2}{*}{client 5} & FedGAIRAT & 0.00 & 0.00 & 9.78 & 0.00 & 97.60 & 8.63 & 0.00 & 0.00 & 0.00 & 0.00 & 11.60 \\
 &  & CalFAT(ours) & 44.55 & 41.50 & 28.10 & 4.33 & 11.18 & 29.74 & 17.07 & 21.32 & 25.71 & 20.35 & 24.39 \\
 \bottomrule
\end{tabular}}
\end{table}

\subsection{Per-class Average Performance}
\label{appsec:per_class_avg}
\begin{figure}[H]
     \centering
     \subfigure{
		\includegraphics[width=0.4\linewidth]{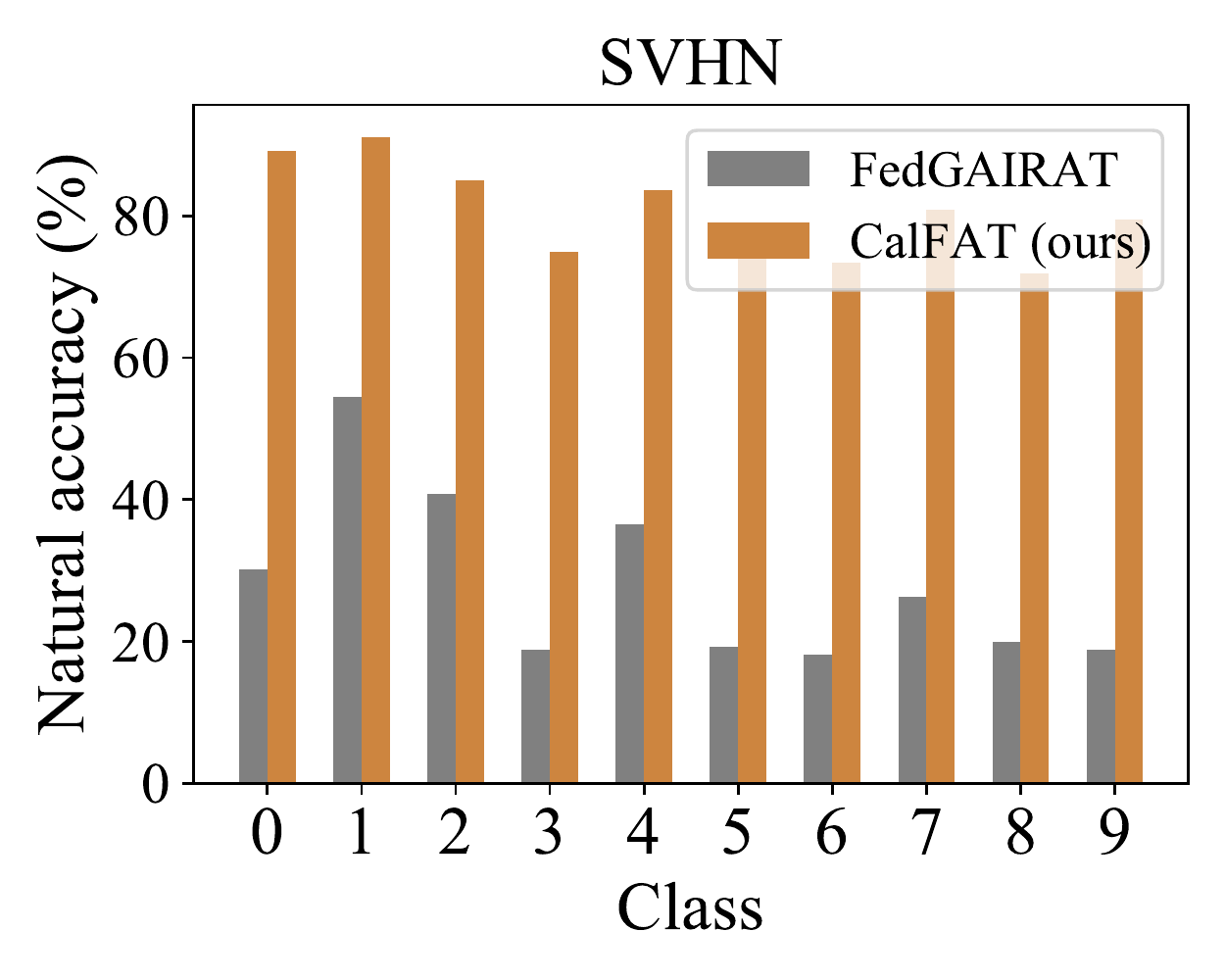}
	}
	\hspace{-4mm}
	\subfigure{
		\includegraphics[width=0.4\linewidth]{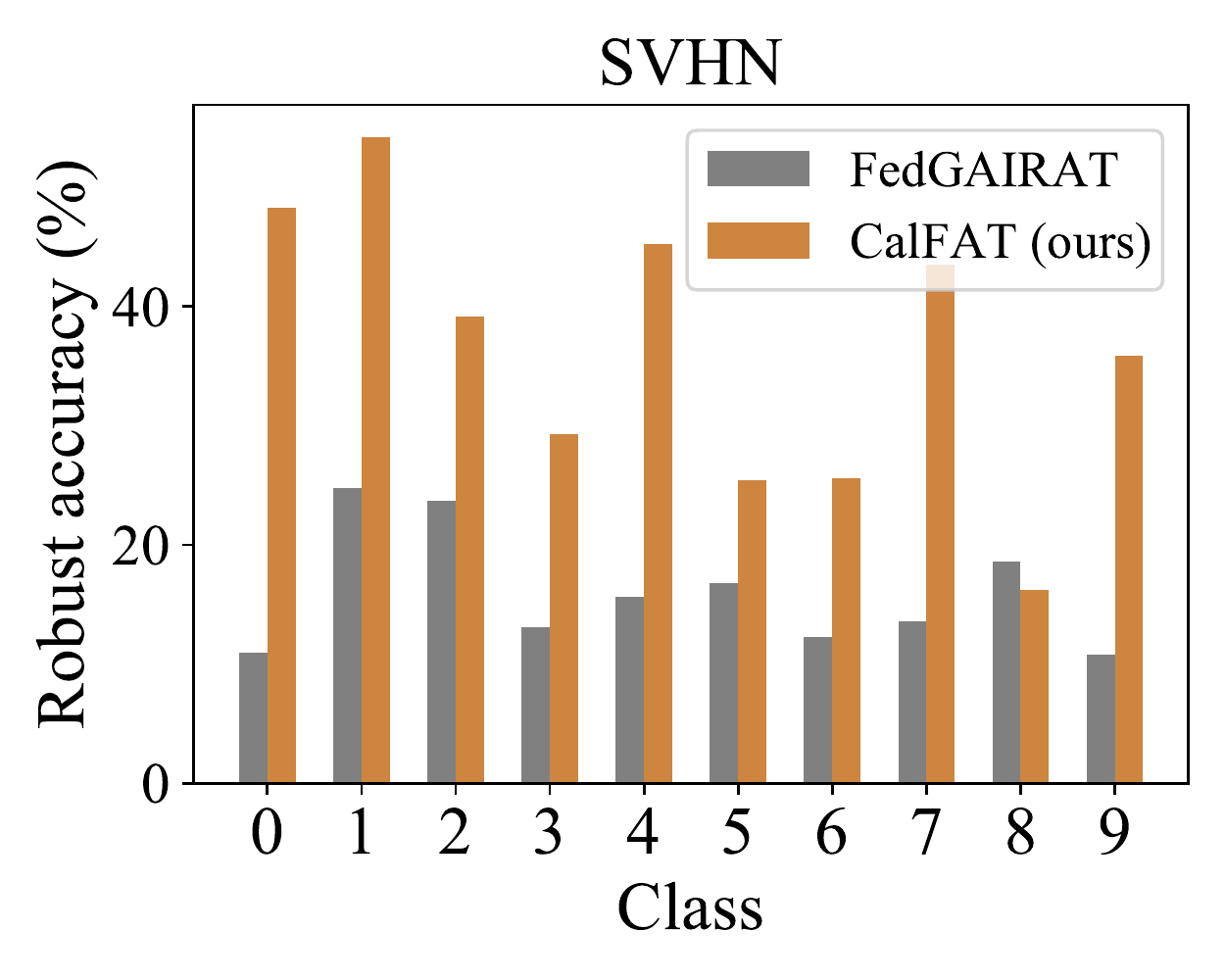}
	}
     \caption{Per-class natural accuracy and robust accuracy (against PGD-20 attack) of CalFAT and the best baseline (FedGAIRAT) on SVHN dataset.}
     \label{appfig:perclass_svhn}
\end{figure}

Figure~\ref{appfig:perclass_svhn} shows the per-class average performance on SVHN dataset.

\subsection{Evaluation on Different Network Architectures}
\label{appsec:arch}
Table~\ref{apptbl:architecture} shows the natural and robust accuracies with different network architectures on CIFAR10 dataset.
\begin{table}[h]
\caption{Natural and robust accuracies (\%) with different network architectures on CIFAR10 dataset. 
}
\label{apptbl:architecture}
\centering
\scalebox{1}{
\begin{tabular}{c|cc|cc|cc}
\toprule
Network & \multicolumn{2}{c|}{CNN} & \multicolumn{2}{c|}{VGG-8} & \multicolumn{2}{c}{ResNet-18} \\
\midrule
Metric & Natural & PGD-20 & Natural & PGD-20 & Natural & PGD-20 \\
\midrule
MixFAT & 53.23 & 26.22 & 59.60 & 34.99 & 67.54 & 38.25 \\
FedPGD & 47.21 & 26.50& 62.21 & 34.89 & 65.48 & 30.04 \\
FedTRADES & 46.14 & 26.29 & 47.21 & 30.39 & 54.61 & 35.03 \\
FedMART & 25.68 & 18.15 & 43.28 & 30.16 & 52.13 & 33.24 \\
FedGAIRAT & 48.34 & 27.32 & 47.83 & 30.52 & 55.62 & 34.87 \\
FedRBN & 47.87 & 26.21 & 46.96 & 30.21 & 54.32 & 33.23 \\
\midrule
CalFAT(ours) & \textbf{64.85} & \textbf{31.19} & \textbf{75.05} & \textbf{40.09} & \textbf{76.73} & \textbf{47.85} \\
\bottomrule
\end{tabular}}
\end{table}


\subsection{Visualization of Different Methods}
\label{appsec:tsne}
\begin{figure}[h!]
	\centering
	\subfigure[FedTRADES]{
		\includegraphics[width=0.44\linewidth]{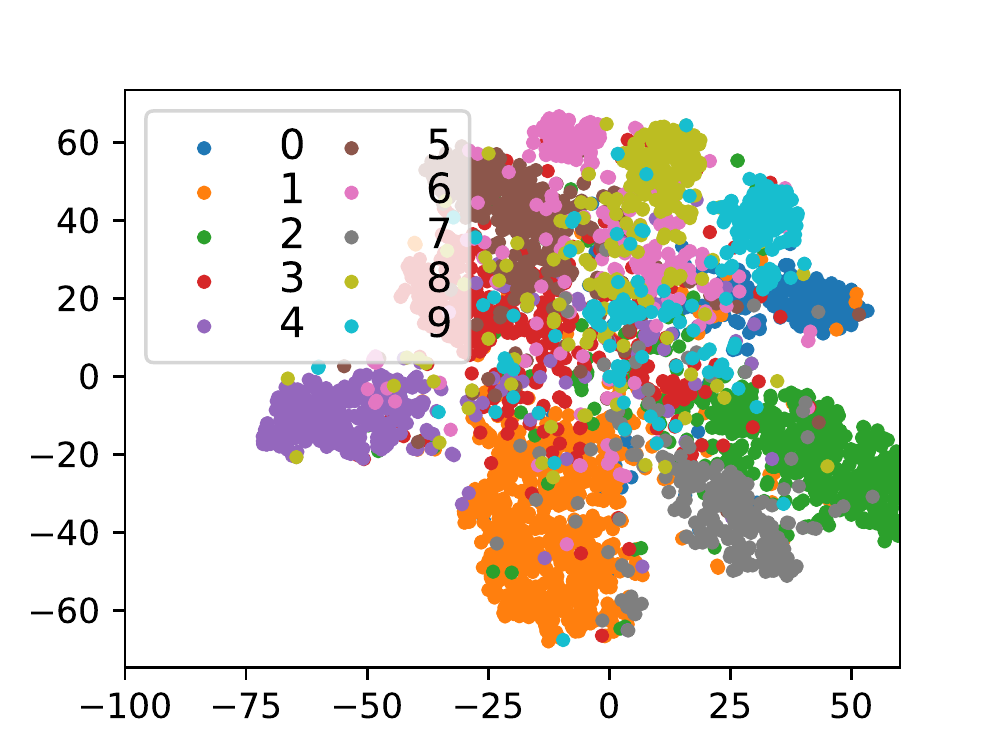}
		\label{subfig:tsne_TRADES}
	}
	\hspace{-7mm}
	\subfigure[CalFAT (ours)]{
		\includegraphics[width=0.44\linewidth]{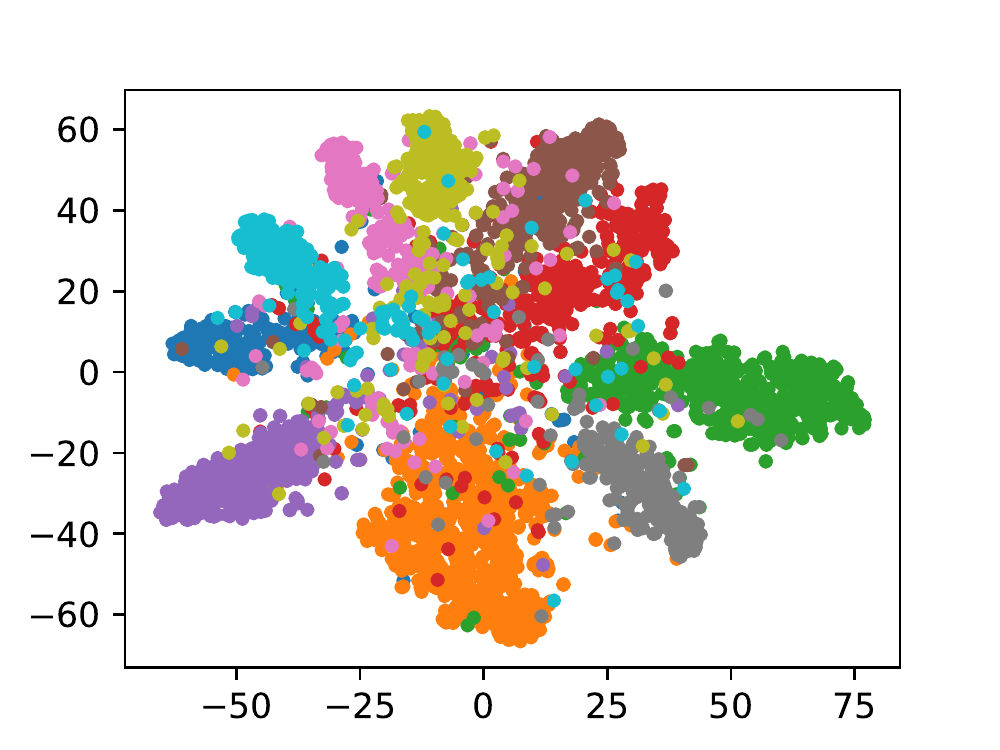}
		\label{subfig:tsne_CalFAT}
	}
	\caption{t-SNE feature visualization of FedTRADES and CalFAT on SVHN dataset. Each color represents a class. Samples from different classes 
	are hard to be separated in FedTRADES while CalFAT can learn more discriminative features.}
	\label{appfig:tsne}
\end{figure}

Figure~\ref{appfig:tsne} shows the t-SNE feature visualization of FedTRADES and CalFAT on SVHN dataset.

\subsection{Performance under the IID setting}
\label{appsec:iid}
\revised{Table \ref{apptbl:iid} shows the natural accuracy and robust accuracy (against PGD-20 attack) on CIFAR10 dataset under the IID setting.}
\begin{table}[H]
\caption{Natural and robust accuracy (\%) on CIFAR10 dataset under the IID setting.
}
\label{apptbl:iid}
\centering
\scalebox{1}{
\begin{tabular}{c|cc}
\toprule
Metric & Natural & PGD-20 \\
\midrule
MixFAT & 79.62 & 37.57 \\
FedPGD & 75.89 & 42.16 \\
FedTRADES & 74.29 & 44.35 \\
CalFAT & 74.23 & 44.68 \\
\bottomrule
\end{tabular}}
\end{table}

\subsection{Impact of the Number of Clients}
\label{appsec:client_number}
Table~\ref{apptbl:clients} shows the natural and robust accuracies with different numbers of clients on CIFAR10 dataset.
\begin{table}[H]
\caption{Natural and robust accuracies (\%) with different numbers of clients $m=\{20,50,100\}$ on CIFAR10 dataset. 
}
\label{apptbl:clients}
\centering
\resizebox{\textwidth}{!}{
\begin{tabular}{c|ccc|ccc|ccc}
\toprule
$m$ & \multicolumn{3}{c|}{20} & \multicolumn{3}{c|}{50} & \multicolumn{3}{c}{100} \\
\midrule
Metric & Natural & PGD-20 & AA & Natural & PGD-20 & AA & Natural & PGD-20 & AA \\
\midrule
MixFAT & 26.59	$\pm$ 0.16 & 18.24	$\pm$ 0.07 & 13.12	$\pm$ 0.14 & 23.28	$\pm$ 0.16 & 15.55	$\pm$ 0.13 & 10.92	$\pm$ 0.14 & 20.85	$\pm$ 0.16 & 14.41	$\pm$ 0.11 & 10.66	$\pm$ 0.12 \\
FedPGD & 29.38	$\pm$ 0.20 & 18.19	$\pm$ 0.18 & 14.22	$\pm$ 0.11 & 27.73	$\pm$ 0.15 & 16.98	$\pm$ 0.23 & 11.94	$\pm$ 0.14 & 23.86	$\pm$ 0.18 & 15.37	$\pm$ 0.18 & 10.78	$\pm$ 0.09 \\
FedTRADES & 29.39	$\pm$ 0.14 & 18.47	$\pm$ 0.13 & 14.66	$\pm$ 0.19 & 21.44	$\pm$ 0.06 & 15.20	$\pm$ 0.16 & 11.85	$\pm$ 0.09 & 21.06	$\pm$ 0.11 & 14.76	$\pm$ 0.16 & 11.68	$\pm$ 0.07 \\
FedMART & 22.95	$\pm$ 0.15 & 17.08	$\pm$ 0.07 & 13.34	$\pm$ 0.09 & 22.43	$\pm$ 0.15 & 15.01	$\pm$ 0.08 & 11.59	$\pm$ 0.06 & 21.58	$\pm$ 0.12 & 14.48	$\pm$ 0.17 & 11.01	$\pm$ 0.09 \\
FedGAIRAT & 22.74	$\pm$ 0.13 & 17.00	$\pm$ 0.12 & 13.77	$\pm$ 0.17 & 20.84	$\pm$ 0.26 & 14.68	$\pm$ 0.21 & 11.80	$\pm$ 0.17 & 19.26	$\pm$ 0.15 & 14.17	$\pm$ 0.11 & 11.33	$\pm$ 0.14 \\
FedRBN & 21.90	$\pm$ 0.13 & 17.46	$\pm$ 0.14 & 12.91	$\pm$ 0.11 & 20.22	$\pm$ 0.16 & 14.74	$\pm$ 0.16 & 12.13	$\pm$ 0.11 & 18.99	$\pm$ 0.11 & 13.48	$\pm$ 0.19 & 12.05	$\pm$ 0.08 \\
CalFAT & \textbf{60.26}	$\pm$ 0.09 & \textbf{24.32}	$\pm$ 0.13 & \textbf{15.41}	$\pm$ 0.12 & \textbf{49.86}	$\pm$ 0.07 & \textbf{18.79}	$\pm$ 0.10 & \textbf{13.22}	$\pm$ 0.13 & \textbf{40.6}9	$\pm$ 0.08 & \textbf{16.19}	$\pm$ 0.15 & \textbf{12.51}	$\pm$ 0.09 \\
\bottomrule
\end{tabular}}
\end{table}

\subsection{Impact of Skewed Label Distribution}
\label{appsec:skew_level}
Table~\ref{apptbl:beta} shows the natural and robust accuracies under different level of label skewness on CIFAR10 dataset.

\begin{table}[H]
\caption{Natural and robust accuracies (\%) under different label skewness levels $\beta$ on CIFAR10 dataset. 
}
\label{apptbl:beta}
\centering
\resizebox{\textwidth}{!}{
\begin{tabular}{c|cccccc|cccccc|cccccc}
\toprule
Label skewness level & \multicolumn{6}{c|}{$\beta=0.05$} & \multicolumn{6}{c|}{$\beta=0.2$} & \multicolumn{6}{c}{$\beta=0.3$} \\
\midrule
Metric & Natural & FGSM & BIM & CW & PGD-20 & AA & Natural & FGSM & BIM & CW & PGD-20 & AA & Natural & FGSM & BIM & CW & PGD-20 & AA \\
\midrule
MixFAT & 49.10 & 27.49 & 25.32 & 22.17 & 25.24 & 22.51 & 54.85 & 31.27 & 28.70 & 26.08 & 28.46 & 25.21 & 58.93 & 31.68 & 28.17 & 24.96 & 28.00 & 24.34 \\
FedPGD & 47.13 & 26.63 & 24.96 & 20.75 & 25.03 & 21.28 & 52.22 & 30.31 & 28.64 & 25.49 & 28.59 & 24.92 & 56.12 & 30.86 & 28.46 & 25.07 & 28.29 & 23.64 \\
FedTRADES & 40.24 & 26.02 & 25.06 & 22.48 & 24.99 & 20.16 & 48.52 & 29.94 & 28.73 & 25.57 & 28.65 & 24.15 & 54.26 & 30.83 & 29.39 & 24.74 & 29.26 & 23.87 \\
FedMART & 29.84 & 21.90 & 21.39 & 18.31 & 21.41 & 17.89 & 38.38 & 27.59 & 27.05 & 23.31 & 26.99 & 21.89 & 40.96 & 28.32 & 27.88 & 23.12 & 27.80 & 22.16 \\
FedGAIRAT & 50.41 & 28.89 & 26.30 & 22.66 & 26.34 & 23.81 & 56.11 & 32.99 & 29.90 & 27.10 & 28.97 & 25.97 & 60.63 & 33.31 & 30.12 & 25.50 & 29.67 & 24.75 \\
FedRBN & 39.35 & 25.92 & 24.40 & 21.55 & 24.77 & 19.47 & 48.42 & 29.59 & 27.74 & 24.67 & 27.86 & 23.78 & 53.54 & 29.88 & 28.76 & 24.11 & 28.63 & 23.14 \\
\midrule
CalFAT(ours) & \textbf{61.00} & \textbf{32.40} & \textbf{29.75} & \textbf{23.55} & \textbf{29.50} & \textbf{25.66} & \textbf{71.55} & \textbf{33.80} & \textbf{30.70} & \textbf{27.25} & \textbf{29.35} & \textbf{26.32} & \textbf{69.95} & \textbf{34.25} & \textbf{30.80} & \textbf{27.76} & \textbf{30.96} & \textbf{26.84} \\
\bottomrule
\end{tabular}}
\end{table}

\subsection{Contribution of the Calibrated Loss Functions}
\label{appsec:loss_contribution}
Table \ref{apptbl:loss_contribution} shows the results of different loss functions.
\begin{table}[h]
\caption{Natural and robust accuracy (\%) of different loss functions. 
}
\label{apptbl:loss_contribution}
\centering
\resizebox{\textwidth}{!}{
\begin{tabular}{c|ccc|ccc|ccc}
\toprule
Label skewness level & \multicolumn{3}{c|}{$\beta=0.05$} & \multicolumn{3}{c|}{$\beta=0.2$} & \multicolumn{3}{c}{$\beta=0.3$} \\
\midrule
Metric & Natural & PGD-20 & AA & Natural & PGD-20  &AA & Natural & PGD-20 &AA\\
\midrule
w/o $\ell_{cce}(\cdot,\cdot,\cdot)$ & 52.59 & 20.55 & 16.37 & 63.49 & 20.37 & 17.83 & 61.61 & 22.42 & 18.24\\
w/o $\ell_{ckl}(\cdot,\cdot,\cdot)$ & 60.05 & 27.59 & 19.79  & 70.60 & 27.18 & 21.94 & 68.27 & 28.56 & 22.68 \\
CalFAT(ours) & \textbf{61.03} & \textbf{29.49}& \textbf{20.35} & \textbf{71.54} & \textbf{29.36} & \textbf{22.96} & \textbf{69.98} & \textbf{30.98} & \textbf{23.21}\\
\bottomrule
\end{tabular}}
\end{table}

\subsection{Impact of the Ratio of Adversarial Data}
\label{appsec:portion}
Table~\ref{apptbl:portion} shows the robust accuracy (against PGD-20 attack) of CalFAT with different ratios of adversarial data.
\begin{table}[h]
\caption{Robust accuracy (\%) of our CalFAT against PGD-20 attack with different ratios of adversarial data. 
}
\label{apptbl:portion}
\centering
\scalebox{1}{
\begin{tabular}{c|ccccc}
\toprule
Ratio ($r$) & 0 & 0.3 & 0.5 & 0.8 & 1 \\
\midrule
SVHN & 1.25 & 32.35 & 37.31 & 38.59 & \textbf{41.64} \\
CIFAR10 & 3.47 & 15.47 & 21.08 & 25.87 & \textbf{31.19} \\
CIFAR100 & 2.60 & 11.08 & 12.19 & 13.01 & \textbf{15.39} \\
\bottomrule
\end{tabular}}
\end{table}

\end{document}